\documentclass[sigconf]{acmart}

\AtBeginDocument{%
  \providecommand\BibTeX{{%
    \normalfont B\kern-0.5em{\scshape i\kern-0.25em b}\kern-0.8em\TeX}}}

\usepackage{algorithmic}
\usepackage[ruled, linesnumbered]{algorithm2e}
\usepackage{subfigure}
\usepackage{multirow}
\usepackage{enumerate}
\usepackage{todonotes}
\usepackage{pifont}
\usepackage{enumitem}

\usepackage{amsmath}

\usepackage{listings}
\usepackage{xcolor}
\usepackage{colortbl}
\usepackage{tikz}
\usepackage{tcolorbox}

\usepackage{titletoc}

\begin{document}

% \setlength{\parskip}{0.1em}

% \title{Make Your Home Safe: Context-aware Unsupervised User Behavior Anomaly Detection in Smart Homes}

% \title{\schemename: Semantic-aware Graph-guided Behavior Sequences Generation with Large Language Models for Smart Homes}

\title{\textsf{SmartBench}: Evaluating LLMs in Smart Homes with Anomalous Device States and Behavioral Contexts}


% \author{Anonymous author(s)}

\author{Qingsong Zou}
\authornotemark[1]
\affiliation{%
  \institution{Tsinghua Shenzhen International Graduate School}
  \city{Shenzhen}
  \country{China}}
\email{zouqs21@mails.tsinghua.edu.cn}

\author{Zhi Yan}
\affiliation{%
  \institution{Jilin University}
  \city{Jilin}
  \country{China}}
\email{yanzhi2422@mails.jlu.edu.cn}
\authornote{The first two authors have equal contribution.}

\author{Zhiyao Xu}
\affiliation{%
  \institution{Tsinghua Shenzhen International Graduate School}
  \city{Shenzhen}
  \country{China}}
\email{xu-zy25@mails.tsinghua.edu.cn}

\author{Kuofeng Gao}
\affiliation{%
  \institution{Tsinghua Shenzhen International Graduate School}
  \city{Shenzhen}
  \country{China}}
\email{gkf24@mails.tsinghua.edu.cn}

\author{Jingyu Xiao}
\authornote{Jingyu Xiao is the corresponding author.}
\affiliation{%
  \institution{The Chinese University of Hong Kong}
  \city{Hong Kong}
  \country{China}}
\email{jyxiao@link.cuhk.edu.hk}

\author{Yong Jiang}
\affiliation{%
  \institution{Tsinghua Shenzhen International Graduate School}
  \institution{Pengcheng Laboratory}
  \city{Shenzhen}
  \country{China}}
\email{jiangy@sz.tsinghua.edu.cn}

\renewcommand{\shortauthors}{Qingsong Zou et al.}



\begin{abstract}
Due to the strong context-awareness capabilities demonstrated by large language models (LLMs), recent research has begun exploring their integration into smart home assistants to help users manage and adjust their living environments. While LLMs have been shown to effectively understand user needs and provide appropriate responses, most existing studies primarily focus on interpreting and executing user behaviors or instructions.

However, a critical function of smart home assistants is the ability to detect when the home environment is in an anomalous state. This involves two key requirements: the LLM must accurately determine whether an anomalous condition is present, and provide either a clear explanation or actionable suggestions.

To enhance the anomaly detection capabilities of next-generation LLM-based smart home assistants, we introduce \textsf{SmartBench}, which is the first smart home dataset designed for LLMs, containing both normal and anomalous device states as well as normal and anomalous device state transition contexts. We evaluate 13 mainstream LLMs on this benchmark. The experimental results show that most state-of-the-art models cannot achieve good anomaly detection performance. For example, Claude-Sonnet-4.5 achieves only 66.1\% detection accuracy on context-independent anomaly categories, and performs even worse on context-dependent anomalies, with an accuracy of only 57.8\%.
More experimental results suggest that next-generation LLM-based smart home assistants are still far from being able to effectively detect and handle anomalous conditions in the smart home environment.
Our dataset is publicly available at \url{https://github.com/horizonsinzqs/SmartBench}.

\end{abstract}

% Due to the 
% some methods are proposd to xxx. However, temporal context, unbalance user behavior and noise behavior bring huge challenges for user modeling in smart homes. 





% \keywords{User Behavior Modeling, Anomaly Detection, Self-supervised Learning.}



\maketitle

\section{Introduction}
\label{sec:intro}
Smart home assistants play an increasingly important role in modern residential life. By leveraging user habits, historical interactions, and physical parameters of the home, these assistants enable automated control of smart home devices (i.e. IoT devices) or provide actionable recommendations, helping users better regulate and manage their living environment~\cite{xiao2023know, ContexloT, jordan2025plant, xiao2023user}.
Notably, anomaly conditions within a smart home can arise from various factors such as user misoperation or device malfunction.  
In addition, the rapid growth in the diversity and adoption of IoT devices has also made them attractive targets for attackers~\cite{zou2023iotbeholder, homole, delay-sp, ozmen2022discovering}. As a result, detecting anomaly conditions within the home environment has become a critical task for smart home systems~\cite{kdd, Rieger23ARGUS, iotgaze, hawatcher, Siker19Aegis, dsnGraph, zhao2025security}.

Due to the strong understanding and generalization capabilities demonstrated by large language models (LLMs) across various domains, many recent studies have focused on integrating LLMs into smart home assistants to provide more intelligent and personalized services for smart home users~\cite{gao2024chatiot, SimuHome, king2023get, xu2025IoTM, xu2025semantic}. For example, King et al.~\cite{king2024sasha} introduced Sasha, which leverages LLMs to translate users’ loosely expressed instructions into personalized device operations. Yu et al.~\cite{yu2026leveraging} proposed IoTGPT, an LLM-based smart home agent designed to execute IoT commands in a reliable, efficient, and personalized manner. Gao et al.~\cite{gao2024chatiot} developed a no-code smart home system powered by LLMs, which automatically generates trigger-action programs for IoT devices based on user requests.
\begin{figure}[t]
\centering %图片全局居中
\subfigtopskip=1pt
\subfigure[Conflicting climate devices]{
\label{subfig:fig1a}
\includegraphics[width=0.48\linewidth]{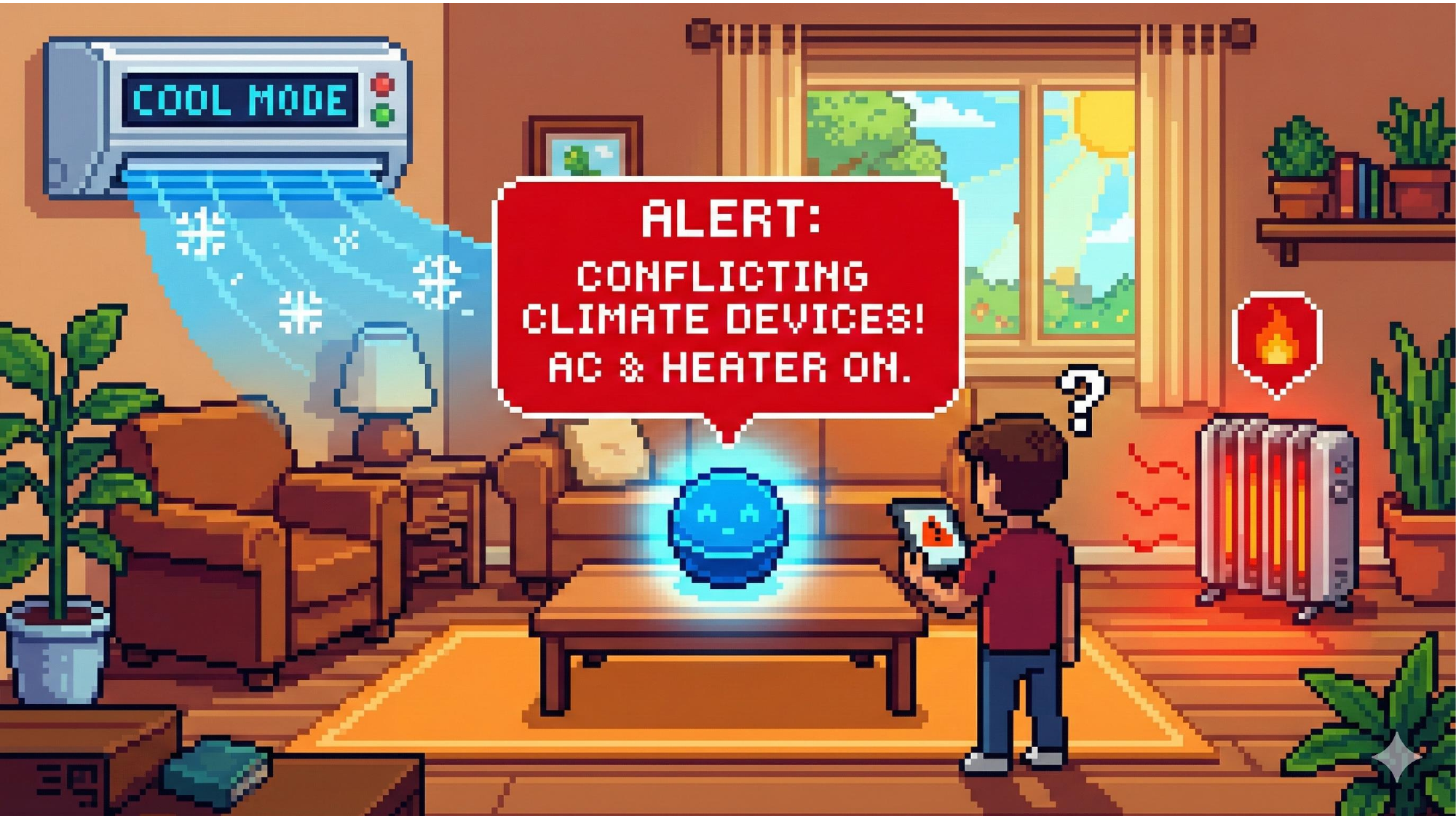}}
\subfigure[Unsafe device state]{
\label{subfig:fig1b}
\includegraphics[width=0.48\linewidth]{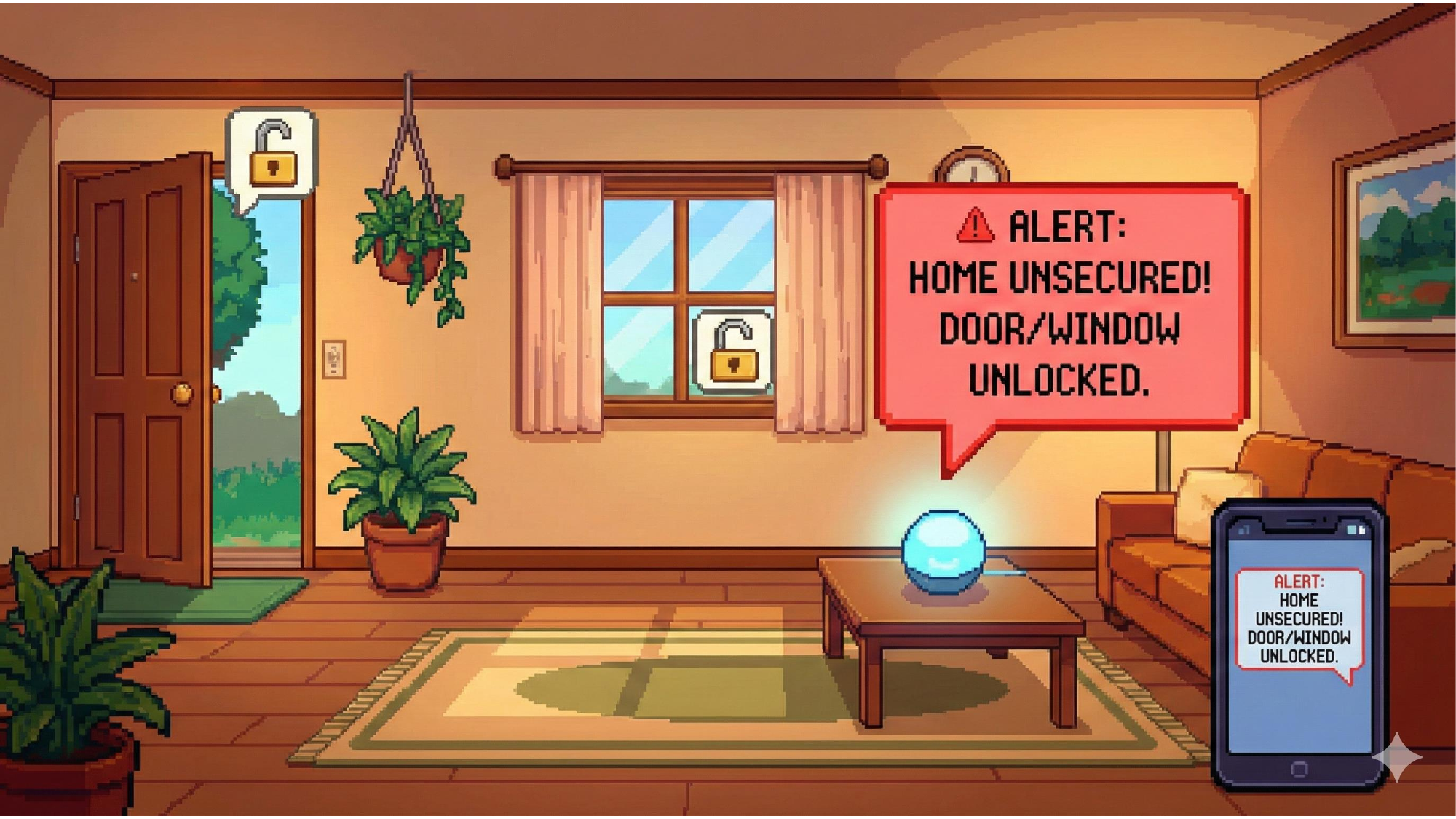}}

\subfigure[Device left running for a long time]{
\label{subfig:fig1c}
\includegraphics[width=0.48\linewidth]{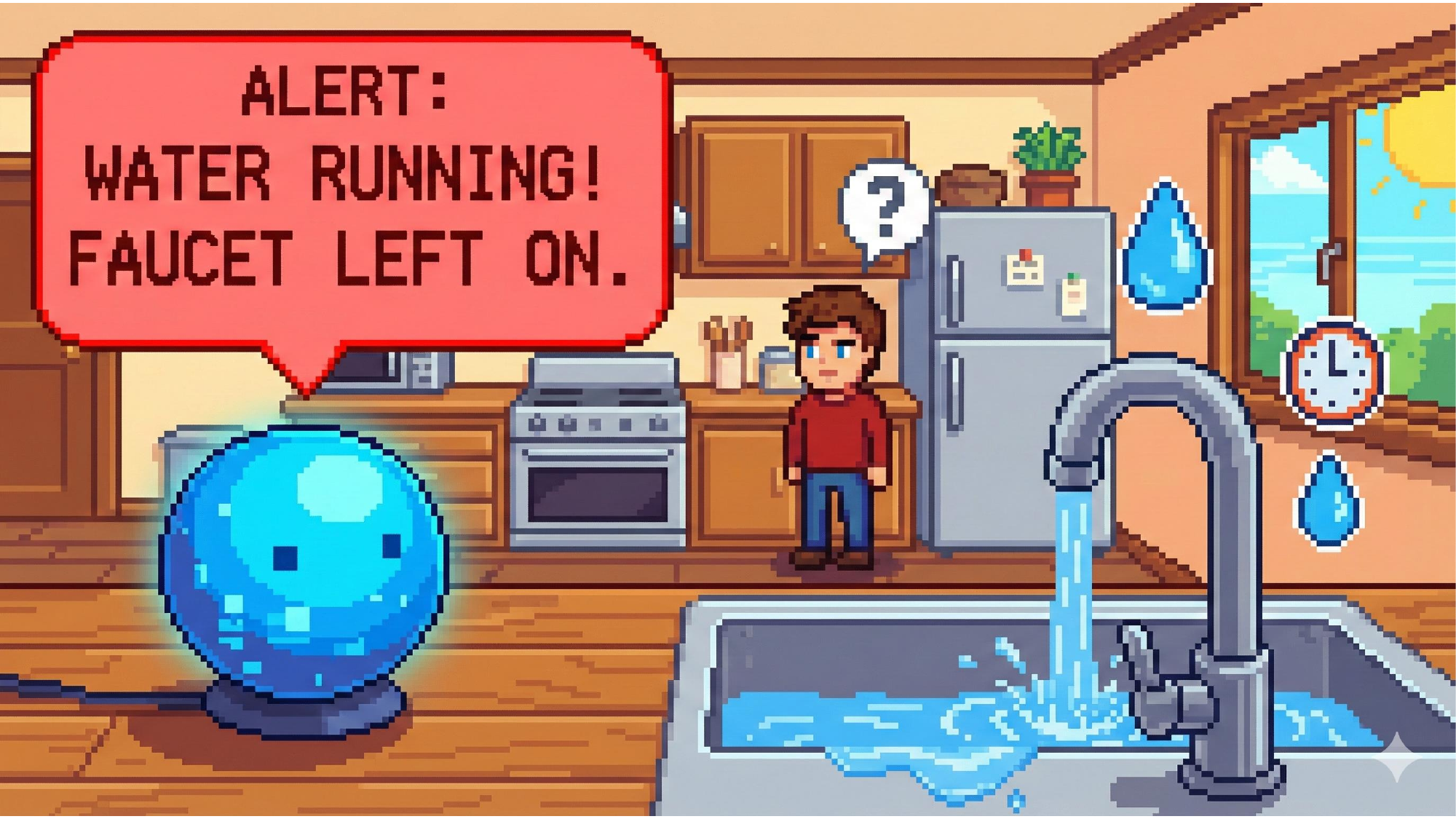}}
\subfigure[Device malfunction]{
\label{subfig:fig1d}
\includegraphics[width=0.48\linewidth]{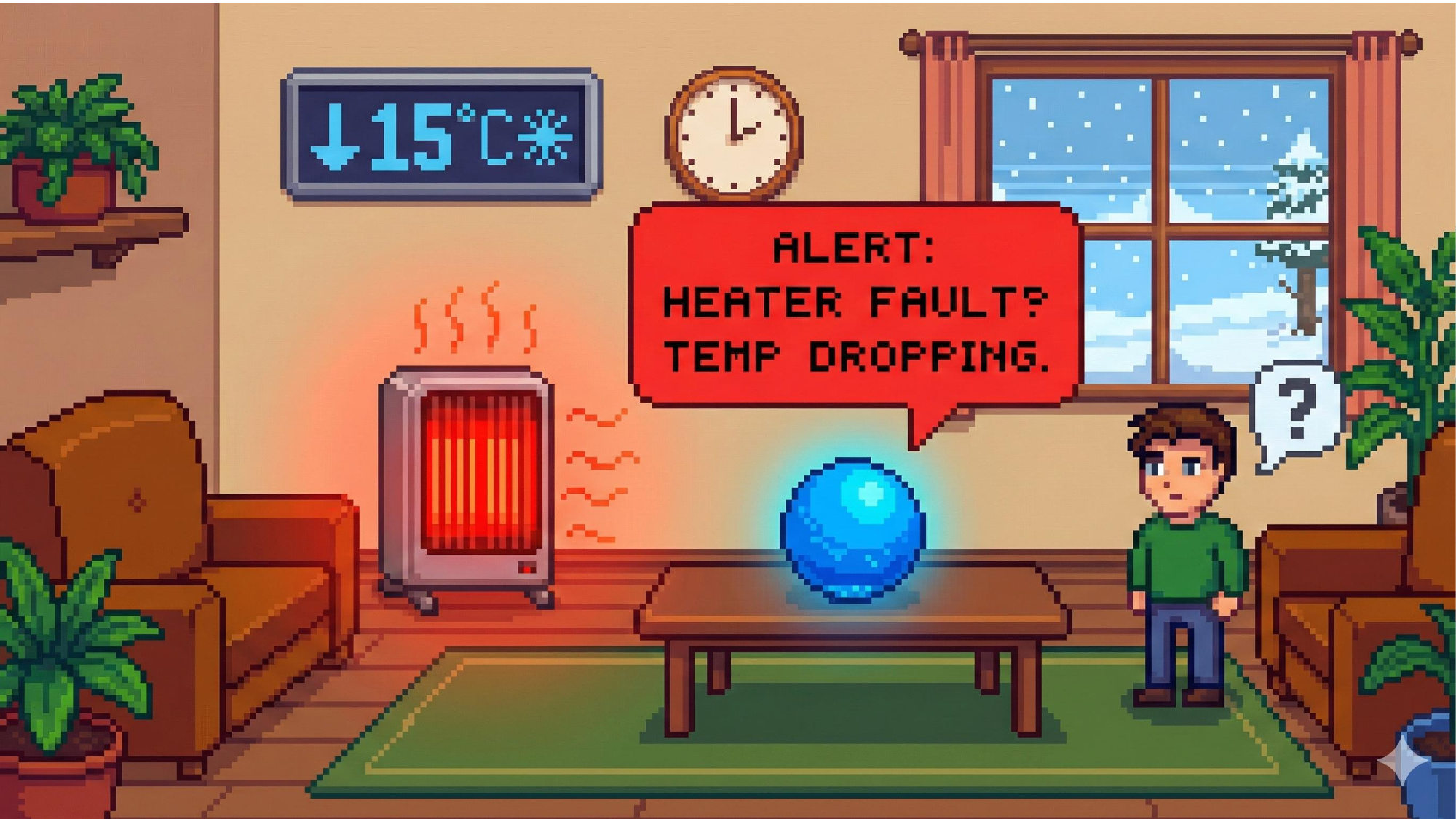}}
\caption{Examples of anomaly conditions in a smart home.}
\label{fig:intro}
\end{figure}

% These studies demonstrate the great potential of LLMs in this field. However, they generally focus on understanding and processing user behaviors or commands, while overlooking a critical capability of smart home assistants: the ability to perceive anomaly conditions within the home environment. In real-world scenarios, there are various potential indicators that may suggest a smart home is in an anomalous state. As illustrated in Figure~\ref{subfig:fig1a}, when the air conditioner is set to cooling mode while the heater is simultaneously turned on, this may result from user error or indicate that the IoT devices are under a power-draining attack. In such cases, the smart home assistant should alert the user and suggest which device to turn off or adjust based on the current indoor temperature.
% Figure~\ref{subfig:fig1b} shows another example. If the smart door lock or window lock is found to be unlocked while the user is away from home, the assistant should notify the user of a potential security risk.
% Similarly, if the kitchen faucet remains on for an extended period (Figure~\ref{subfig:fig1c}), or the humidifier has been running for some time but the humidity sensor shows a continuous drop (Figure~\ref{subfig:fig1d}), the assistant should inform the user of possible misoperation or device malfunction.
These studies highlight the strong potential of LLMs in this domain. However, most focus on interpreting user behaviors or commands and overlook a core capability of smart home assistants: perceiving anomalous conditions in the home environment. In practice, many signals may indicate an anomalous state. As shown in Figure~\ref{subfig:fig1a}, an air conditioner set to cooling while the heater is on may reflect user error or a power-draining attack; the assistant should warn the user and recommend which device to adjust based on the indoor temperature. Figure~\ref{subfig:fig1b} presents another case. If a smart door or window lock is unlocked while the user is away, the assistant should notify the user of a security risk. Likewise, if the kitchen faucet is left running for an extended period (Figure~\ref{subfig:fig1c}), or a humidifier runs while the humidity sensor continuously drops (Figure~\ref{subfig:fig1d}), the assistant should alert the user to potential misoperation or device malfunction.

% Identifying these anomalies is not straightforward, as their causes are diverse and require careful consideration of multiple details. For example, the anomalies shown in Figures~\ref{subfig:fig1a} and ~\ref{subfig:fig1b} can be detected by inspecting device states or readings at a specific point in time. However, the issues depicted in Figures ~\ref{subfig:fig1c} and ~\ref{subfig:fig1d} require monitoring the states or readings of one or more devices over an extended period.
% Moreover, smart home users often expect not only to be alerted when an anomaly occurs, but also to receive actionable suggestions or, at least, an explanation or localization of the anomaly~\cite{kdd, hawatcher}. This implies that simply detecting the presence of an anomaly is insufficient for a smart home assistant.
Identifying these anomalies is challenging because their causes vary and often require jointly considering multiple details. For instance, the anomalies in Figures~\ref{subfig:fig1a} and \ref{subfig:fig1b} can be detected from device states  at a single time point, whereas those in Figures~\ref{subfig:fig1c} and \ref{subfig:fig1d} require tracking one or more devices over time. Moreover, users typically expect not only alerts but also actionable recommendations, or at least a localization of the anomaly~\cite{kdd, hawatcher}. Thus, simply detecting the presence of an anomaly is insufficient. The diversity of IoT devices and application scenarios further complicates detection, explanation, and attribution. Consequently, enabling LLMs to identify smart home anomalies and produce accurate explanations is critical for advancing smart home systems.

To bridge this gap, we introduce a new comprehensive evaluation benchmark: \textsf{SmartBench}, which is the first smart home dataset designed for LLMs, containing both normal and anomalous device states as well as normal and anomalous device state transition contexts. Specifically, we construct two types of in-home anomalies: \textit{context-independent} and \textit{context-dependent}, covering 15 anomaly categories and totaling 4,400 samples.
Each context-independent sample contains the states or readings of 62 IoT devices at a specific point in time. Each context-dependent sample consists of a sequence of device actions (state transitions) with corresponding timestamps, with sequence lengths ranging from 20 to 4732. All anomalous samples are also annotated with the criteria used to determine the anomaly, which can be used to further evaluate the reliability of LLM-based anomaly detection.
By introducing \textsf{SmartBench}, we aim to provide an effective evaluation tool for developing the next generation of smart home assistants that are capable of accurately understanding, detecting, and dealing with complex anomalous states in smart home environments, thereby ensuring a safer and more reliable experience for users.

As a conclusion, our main contributions are as follows:
\begin{itemize}
[topsep=0pt,itemsep=0pt,parsep=0pt,partopsep=0pt,leftmargin=10pt]
\item To the best of our knowledge, we are the first to investigate the ability of LLMs to detect anomalous states in smart home environments.
\item We propose \textsf{SmartBench}, a benchmark that includes a diverse set of benign and anomalous in-home scenarios. It covers a wide range of anomaly types, including both context-independent and context-dependent, and a rich variety of device actions.
\item Our experiments on 13 mainstream LLMs show that almost all models perform poorly on this benchmark. Further analysis reveals that even when a model is able to recognize the presence of an anomaly, it often fails to correctly analyze the underlying cause. This indicates that LLMs are still a long way from achieving reliable detection and handling of anomalous conditions in home environments.
\end{itemize}

\section{Related Work}
\label{sec:relatedwork}

\subsection{Smart Home Assistant on LLM}

A smart home assistant is expected to support multiple capabilities simultaneously, including understanding and executing user commands, perceiving the indoor environment and detecting anomalies, and modeling user behaviors to provide personalized suggestions. Traditional machine learning and deep learning methods often struggle to support such multi-task requirements.

Motivated by the strong performance of LLMs across domains, recent work has explored incorporating LLMs into smart home assistants to handle complex tasks in a unified manner~\cite{xiaomi, king2024sasha, king2023get, civitarese2025large, polo2025enhancing, yu2026leveraging}. Civitarese et al.~\cite{civitarese2025large} propose ADL-LLM for recognizing activities of daily living from sensor data. King et al.~\cite{king2024sasha} develop Sasha to interpret complex and ambiguous user instructions. Yu et al.~\cite{yu2026leveraging} introduce IoTGPT for efficient execution of personalized IoT commands. Polo et al.~\cite{polo2025enhancing} present an LLM-based architecture for context-aware interaction in smart environments. These studies demonstrate the potential of LLMs in enhancing the intelligence and automation of smart homes. However, they do not discuss the ability of LLMs to detect anomalous conditions within the smart home environment, leaving a significant gap between current approaches and the expected fully integrated smart home assistant.

\subsection{Anomaly Detection in Smart Homes}
% Anomaly detection has long been a key research topic in smart home systems. In this context, smart home devices serve as both sensors and actuators, which means that these devices can directly alter the physical environment of the home (e.g., changing temperature, humidity, or spatial accessibility, etc.). Anomalous behaviors such as improper user operations~\cite{ozmen2022discovering} or potential attacks by malicious actors can pose serious security risks~\cite{delay-dsn, delay-sp}. As a result, whenever a smart home device's state is changed, it is necessary to assess its impact on the home environment and monitor whether any anomalous conditions arise.

% In previous studies, most approaches have leveraged machine learning or deep learning techniques to analyze and model user interactions with devices, enabling anomaly detection capabilities~\cite{hawatcher, homonit, iotgaze, Rieger23ARGUS, kdd, chi2025iotbystander}.
% For example, Rieger et al.~\cite{Rieger23ARGUS} proposed an autoencoder-based method for anomaly detection in smart homes, enabling the detection of previously unseen anomalies. Xiao et al.~\cite{kdd} introduced SmartGuard, an unsupervised learning framework capable of effectively detecting anomalies from noisy, long-context user behavior sequences.
Anomaly detection has long been a central problem in smart home systems. Because smart home devices act as both sensors and actuators, they can directly affect the physical home environment (e.g., temperature, humidity, and accessibility). Anomalies arising from user misoperations~\cite{ozmen2022discovering} or malicious attacks can therefore create serious security risks~\cite{delay-dsn, delay-sp}. Accordingly, when a device state changes, it is important to assess its environmental impact and monitor for anomalous conditions.

Most prior work uses machine learning or deep learning to model user–device interactions for anomaly detection~\cite{hawatcher, homonit, iotgaze, Rieger23ARGUS, kdd, chi2025iotbystander}. For instance, Rieger et al.~\cite{Rieger23ARGUS} propose an autoencoder-based approach to detect previously unseen anomalies. Xiao et al.~\cite{kdd} introduce SmartGuard, an unsupervised framework that detects anomalies from noisy long-context behavior sequences.
Chi et al.~\cite{chi2025iotbystander} and Gu et al.~\cite{iotgaze} infer user–device interactions from network traffic or wireless signals and then detect anomalies based on these interactions. Although effective, such methods typically stop at detection and rarely provide actionable guidance after an anomaly is found, making them more suitable as auxiliary tools than as an smart assistant.

With the recent push to integrate LLMs into smart home assistants, emerging work has started to explore LLMs for this task~\cite{jamshidi2025role}. For example, Zeng et al.~\cite{zeng2025large} propose an LLM-based IoT security assistant that incorporates CoT to improve understanding of vulnerabilities and threats. Chowdhury et al.~\cite{chowdhury2025identifying} develop a security-focused Q\&A system to help users address common IoT issues, and Dong et al.~\cite{dong2025chatiot} introduce ChatIoT to analyze and disseminate IoT security and threat intelligence.

However, these works either focus on device firmware and network traffic, overlooking the physical environment and human factors within the smart home, or function solely as Q\&A bots without discussing the LLM’s ability to detect and analyze anomalous conditions occurring in real smart home environments.

% \subsection{Benchmark on Smart Homes}
% As more research explores the integration of LLMs into smart home environments, several studies have begun developing high-quality benchmarks to evaluate the performance of LLM-based smart home assistants.
% Seo et al.~\cite{SimuHome} introduced a time-accelerated home environment that simulates smart devices supports API calls, and reflects changes in environmental variables. Jordan et al.~\cite{jordan2025plant} developed a text-based benchmark to assess LLMs’ capabilities in incremental learning (based on environmental feedback) and controlled context adaptation. Li et al.~\cite{homebench} evaluated LLMs’ ability to understand and handle valid, invalid, and mixed user requests in both single-device and multi-device scenarios.

% However, these benchmarks primarily focus on understanding user commands and enabling personalized recommendations, while paying little attention to detecting anomalous conditions within the smart home environment.

\section{SmartBench}
\label{sec:smartbench}

% \subsection{Motivation}
% \label{subsec:motivation}

\subsection{Task Definition}
\label{subsec:taskdefinition}

Assume a smart home environment where a set of IoT devices (denoted as $D = \{d_1, d_2, d_3, \ldots\}$) is deployed within a specific location(denoted as $L = \{ l_1, l_2, l_3, \ldots \}$). 
% Each device provides a set of callable functions $F_d = \{f_1, f_2, f_3, \ldots\}$, and at any given time, 
At any given time, each device is in a corresponding state $S_{d} = \{s_1, s_2, s_3, \ldots\}$ (e.g., a smart lock can be in an "unlocked" or "locked" state depending on whether the unlock or lock function was called).

Additionally, some specific IoT devices or sensors can read physical environmental parameters $E = \{T\ \text{(temperature)}, H\ \text{(humidity)}, \\ AQ\ \text{(air quality)}, L\ \text{(light)}, W\ \text{(weather)}, D\ \text{(day of week)}, \ldots\}$. \\ These parameters can be accessed by the smart home assistant at any time. Typically, indoor environmental parameters readings in a smart home are collected periodically.

For \textbf{context-independent anomalies}, the LLM is required to determine the set of potentially anomalous devices $D_{\text{error}}$ based on the device state snapshot at a specific time, denoted as $S_t = \{s_{d,t}, s_{d,t}, s_{d,t}, \ldots\}, d \in D$, along with the current environmental parameters $E_t = \{T_t, H_t, AQ_t, B_t\}$. In addition, the LLM must provide a clear rationale, denoted as $R$, for identifying each anomaly. The overall input-output mapping is defined as:
\begin{equation}
   [S_t, E_t] \rightarrow D_{\text{error}}, R, 
\end{equation}
if no anomaly is present, the output should simply be \text{normal}.

% For context-dependent anomalies, the LLM must analyze a sequence of device operations and corresponding environmental parameter changes over time to determine whether an anomaly has occurred. 
For \textbf{context-dependent anomalies}, the LLM must analyze the sequence of device state transitions, and the temporal evolution of environmental parameters to determine whether an anomaly has occurred. 
In addition, we include two high-level information $I$: a brief household profile (e.g., single adult office worker, family on vacation, night shift worker, etc.) and the duration of the this segments for the LLM to make more informed decisions. 
Let each environmental parameter at a specific time be treated as a special type of device state $s_{e,t}$, where $e \in E$. Let $s_{x,t}$ denotes the state of device $x$ at time $t$, $l_x$ represents the room where device $x$ is located, and $t$ denote the timestamp.

Given a sequence of state transitions over a time window $A = \{(t_1, x_1, s_{x_1, t_1}, l_x), (t_2, x_2, s_{x_2, t_2}, l_x), (t_3, x_3, s_{x_3, t_3}, l_x), \ldots\}, \text{where} \ x \in \\ D \cup P, \ s_{x,t} \in S_x,$
% $A = \{(f_{x, t_1}, \\ s_{x, t_1}, t_1, l_x), (f_{x, t_2}, s_{x, t_2}, t_2, l_x), (f_{x, t_3}, s_{x, t_3}, t_3, l_x), \ldots\}, \text{where} \ x \in D \cup E, f_{x,t} \in F_x, s_{x,t} \in S_x,$
the LLM is expected to infer the potentially anomalous device set $D_{\text{error}}$ and provide a rationale $R$ for the detected anomaly. The corresponding input-output mapping is:
\begin{equation}
    [A, I] \rightarrow D_{\text{error}}, R,
\end{equation}
if no anomaly is detected, the output should be \text{normal}.
\begin{figure*}[htbp]
\centering
\includegraphics[width=0.95\linewidth]{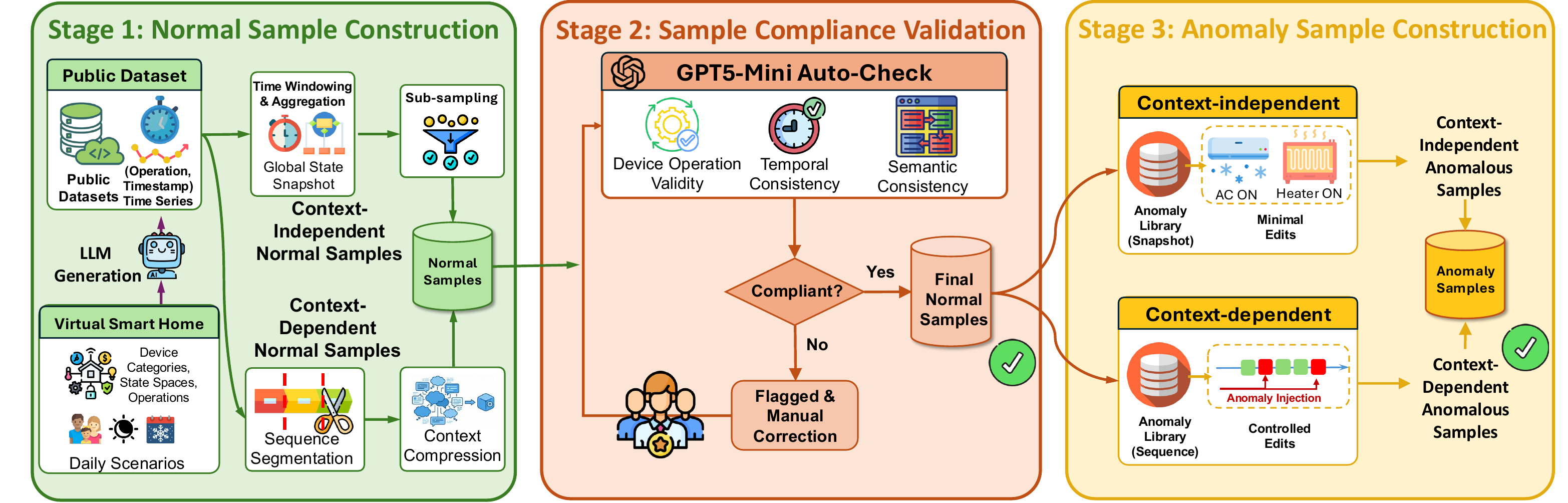}
\caption{Dataset collection process of \textsf{SmartBench}.}
\label{fig:datacollection}
% \vspace{-.8\baselineskip}
\end{figure*}

% \subsection{Anomaly Types}
% \label{subsec:anomalytypes}

\subsection{Dataset Collection}
\label{subsec:datasetcollection}
To systematically evaluate LLM-based smart-home anomaly detection, we define two anomaly types: \textit{context-independent} and \textit{context-dependent}. Context-independent anomalies are snapshots of device states and environmental parameters at a given time, whereas context-dependent anomalies are timestamped sequences of device state transitions, requiring understanding of temporal semantics and causal constraints.

Because collecting data from real smart-home development environments is costly, we build our dataset via a three-stage pipeline: normal sample construction, compliance checking, and anomaly sample construction. Figure~\ref{fig:datacollection} illustrates this process.

\subsubsection{Normal sample construction.} 
We first adopt public smart-home datasets collected from real environments as the source of normal behavior. These datasets must be organized as time series of (\text{timestamp}, \text{device}, \text{state}, \text{location}) tuples. In particular, we use the dataset from~\cite{Rieger23ARGUS} as our primary normal-behavior source.

For context-independent samples, we define a discrete state space for each device type (e.g., an air conditioner can be in cooling, dehumidifying, or off mode) and use a fixed-length time window. At the end of each window, we aggregate all device states and environmental parameters into a global snapshot, which forms one context-independent normal sample.

For context-dependent samples, we split the time series into continuous segments with varying window sizes. We downsample overly long segments by removing states from devices that contribute little contextual semantics. We also attach two high-level descriptors: a brief household profile (e.g., single adult office worker, family on vacation, night shift worker) and the segment duration to support more informed LLM judgments. Each segment plus these descriptors constitutes a context-dependent normal sample.

However, open-source datasets that satisfy our requirements are scarce, so we also use a virtual smart-home environment. We maintain a set of device categories, each with a predefined discrete state space, and use common daily routines or household scenarios as generation units. Each scenario is specified in standardized natural language and then mapped to device state-transition sequences.For example, a “returning home from work” scenario may be described as: a single adult office worker arrives home at around 18:30, unlocks the door, turns on the living room lights, and sequentially activates the air conditioner, humidifier, and television. The entire event takes place within 10 minutes.

We use GPT-5 to generate samples from these scenario descriptions. For context-independent samples, we provide the device list and allowed states and ask GPT-5 to produce the resulting end-of-event device-state snapshot. For context-dependent samples, we prompt GPT-5 to generate (\text{timestamp}, \text{device}, \text{state}, \text{location}) sequences and the accompanying high-level descriptors.

Notably, some context-dependent samples can span long durations and become too lengthy for LLMs. Limited context windows and the quadratic cost of self-attention hinder long-sequence processing, often causing the “lost-in-the-middle” effect and degrading anomaly detection performance~\cite{liu2024lost, huang2024advancing}.

To address this, we design a context compression strategy. Specifically, for the device state transition sequence of context-dependent samples, we merge or remove consecutive, repetitive, and semantically uninformative device state transitions, such as repeated readings of unchanged environmental parameters, gradual air conditioner temperature adjustments, or frequent channel switching on a television. In these cases, we retain only the initial and final states in the sequence and delete other actions. For any sequence of continuous device state transitions, a change in the state of another device or an environmental parameter during that period may alter the semantic interpretation of the original state transitions within the context. Our strategy follows a minimal semantic impact principle, meaning that we ensure that the compressed sequences do not alter the original semantic integrity.

\subsubsection{Sample compliance validation.} 
LLM-generated samples may include hallucinations, leading to unrealistic or logically inconsistent content. For context-independent samples, such errors typically appear as implausible environmental parameter values or references to non-existent devices or states. we address these issues with validation functions that check these elements.
For context-dependent samples, further validation is needed to verify the semantics of device state transition sequences. To ensure each synthetic sample is consistent with its scenario description and free of contextual contradictions, we conduct a compliance-checking procedure for every generated context-dependent sample, based on three criteria:
\begin{enumerate}
[topsep=0pt,itemsep=0pt,parsep=0pt,partopsep=0pt,leftmargin=15pt]
\item Device state transition validity: Each device state transition must match the device type and its defined state space, avoiding non-existent actions or disallowed state transitions.
\item Temporal consistency: Timestamps and their intervals must be reasonable and align with the time description in the scenario.
\item Semantic consistency: The sequence of device state transitions in the sample must align with the corresponding scenario described in natural language.
\end{enumerate}

We leverage the semantic understanding capabilities of LLMs to assist with this validation. Specifically, we use GPT-5-mini to automatically examine each sample for compliance with the above criteria. Any sample that violates one or more criteria is flagged as non-compliant and then will be manually corrected based on suggestions provided by the model. Appendix~\ref{app:compliancevalidation} presents the prompt we used during this process.

After the LLM review, we conducted a manual review of 200 randomly selected synthetic context-dependent samples. The results showed that 197 out of 200 samples (98.5\%) were compliant. Among them, 39 samples were flagged by GPT-5-mini as needing revision. All of these passed the manual review after being corrected. These results demonstrate the effectiveness of GPT-5-mini for compliance checking and confirm that the final dataset contains high-quality samples that realistically reflect smart home scenarios.

\subsubsection{Anomaly sample construction.} 
Collecting anomaly samples in real-world environments is  costly and difficult to scale. For instance, simulating device malfunctions may require firmware modifications, and long-context sequences typically demand extended manual intervention and observation. To address this, we choose to simulate virtual anomalies to represent anomalous conditions.

We first define a category library of context-independent anomalies (i.e., an anomaly pattern library), where each category corresponds to a state combination that have physical/semantic constraints. For each anomaly type, we select normal samples containing the relevant target state variables and apply minimal edits to transform them into anomalies. For example, in the case shown in Figure 1(a), we identify samples that include both air conditioner and heater state variables, and then set both to the "on" state simultaneously to construct a typical conflicting state anomaly.

For context-dependent anomalies, we also begin by defining an anomaly category library and manually author a collection of anomalous sequence fragments for each category. These fragments are then inserted into normal sequences to form anomalous samples. It is important to note that the insertion process follows temporal and semantic consistency constraints, rather than random injection. Each anomaly fragment must be coherent with its preceding context and must not introduce explicit contradictions. For example, after a device failure occurs, its normal executable actions should no longer appear, as this would break narrative consistency. To this end, we define four edit operations for controlled anomaly injection:
\begin{itemize}
[topsep=0pt,itemsep=0pt,parsep=0pt,partopsep=0pt,leftmargin=10pt]
\item Insertion: Introduce a small number of key device state transitions at specific timestamps in the original sequence to explicitly introduce new evidence of anomalies, without altering the main context of the scenario.
\item Deletion: Remove transitions according to the target anomaly mechanism, interrupting an otherwise normal causal chain.
\item Temporal Shift: Keep the order unchanged, but shift key transitions to later timestamps to create temporal inconsistencies.
\item Value Adjustment: Maintain the event structure while modifying key observed values under constraints, thereby introducing abnormal deviations in magnitude, trend, or correlations.
\end{itemize}
Without altering the overall scenario background, we apply these four types of edit operations to a small number of key events in a controlled manner, thereby constructing anomalous samples that are paired one-to-one with normal samples. These anomalies are designed to be indistinguishable from normal behavior without proper context, requiring temporal and semantic reasoning for accurate detection.

\subsection{Data Statistic and Analysis}
\label{subsec:datastatiticandanalysis}
In total, our dataset contains 4,400 samples. Based on whether temporal context is required for detection, we divide the samples into two categories: context-independent and context-dependent. A context-independent sample includes the states of several devices and corresponding environmental parameters at a specific point in time. A context-dependent sample consists of a sequence of (timestamp, device, device state, location) pairs. Examples of the data formats for both types of samples are provided in Appendix~\ref{app:examplesample}. 
\begin{figure}[H]
\centering %图片全局居中
\subfigtopskip=1pt
\subfigure[Context-Independent]{
\label{subfig:anomalytype_CI}
\includegraphics[width=0.48\linewidth]{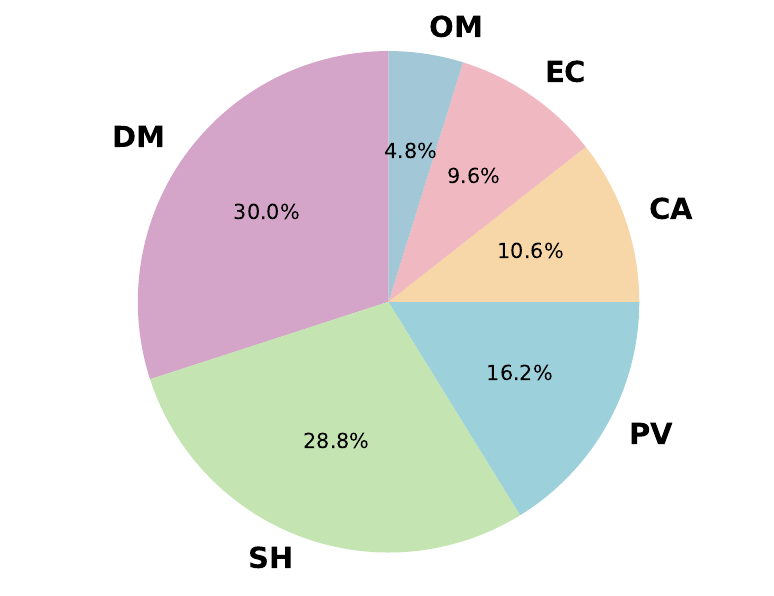}}
\subfigure[Context-Dependent]{
\label{subfig:anomalytype_CD}
\includegraphics[width=0.48\linewidth]{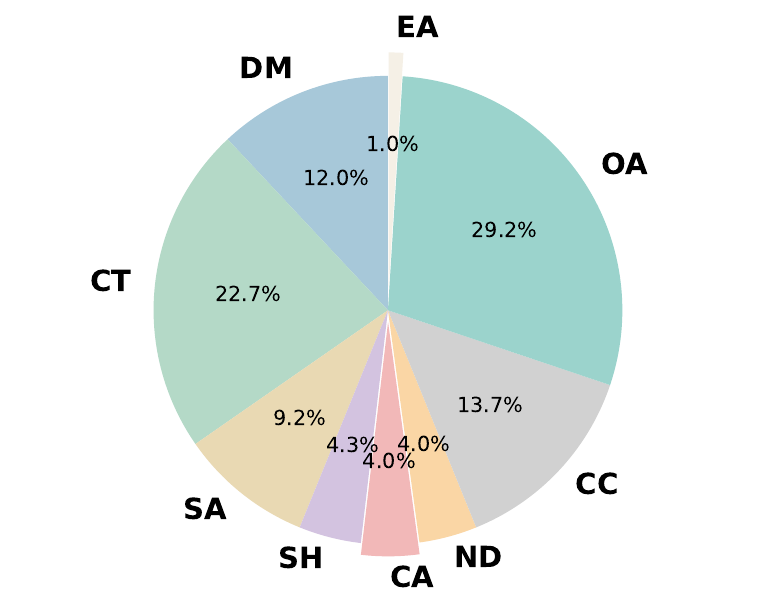}}
\caption{Anomaly types and percentages of \textsf{SmartBench}.}
\label{fig:anomalytype}
\end{figure}

\begin{sloppypar}
To ensure balanced evaluation, we maintain a 1:1 ratio of normal to anomalous samples. During the generation of normal samples by LLMs, we prioritized physical environments and time periods that were underrepresented in the open-source dataset, in order to better balance the overall distribution of the dataset. 
For the context-independent task, we include six types of anomalies:
device\_malfunction (DM), safety\_hazard (SH), physical\_violation (PV), compound\_anomaly (CA), environmental\_conflict (EC), and occupancy\_mismatch (OM).
For the context-dependent task, we include nine types of anomalies:
device\_malfunction (DM), causal\_temporal (CT), statistical\_anomaly (SA), safety\_hazard (SH), climate\_anomaly (CA), network\_disconnect (ND), control\_conflict (CC), occupancy\_anomaly (OA), and energy\_anomaly (EA).
Figure~\ref{fig:anomalytype} presents the context-independent and context-dependent anomaly categories and percentage distribution in \textsf{SmartBench}. 
\end{sloppypar}

We measure the size of context-independent samples by counting the number of characters in each sample. For context-dependent samples, due to their longer context, we quantify size by the number of (\text{timestamp}, \text{device}, \text{state}, \text{location}) tuples, that is, the length of the device state transition sequence in the sample.
General statistics about the dataset are summarized in Table~\ref{tab:statistics} (\#: Count, CI: context-independent sample, CD: context-dependent sample, Anom: anomaly, Atype: anomaly type).
\begin{table}[h]
\centering
\caption{General statistics of \textsf{SmartBench}.}
\label{tab:statistics}
\resizebox{\linewidth}{!}{
\begin{tabular}{cc|cc}
\hline
Statistics   & Description                                                                  & Statistics                                                      & Description                                                                                                                                    \\ \hline
Time of Day  & \begin{tabular}[c]{@{}c@{}}morning, midday, \\ afternoon, night\end{tabular} & Room types                                                      & \begin{tabular}[c]{@{}c@{}}basement, bathroom, bedroom, \\ entrance, garage, hallway,\\ kitchen, living\_room\end{tabular}                     \\ \hline
Season       & \begin{tabular}[c]{@{}c@{}}spring, summer, \\ fall, winter\end{tabular}      & \begin{tabular}[c]{@{}c@{}}Environment\\ Parameter\end{tabular} & \begin{tabular}[c]{@{}c@{}}temperature, humidity, \\ light, motion, weather, \\ smoke, gas, moisture\\ day\_of\_week, is\_weekend\end{tabular} \\ \hline
\#Occupancy  & 29                                                                           & \#Duration                                                      & 15                                                                                                                                             \\ \hline
\#Device     & 155                                                                          & \#State                                                         & 274                                                                                                                                              \\ \hline
\#CI Normal  & 1000                                                                         & \#CD Normal                                                     & 1200                                                                                                                                            \\ \hline
\#CI Anom    & 1000                                                                         & \#CD Anom                                                       & 1200                                                                                                                                            \\ \hline
\#CI Atype   & 50                                                                           & \#CD Atype                                                      & 60                                                                                                                                             \\ \hline
Min CI Size  & 2953                                                                         & Min CD Length                                                   & 20                                                                                                                                             \\ \hline
Max CI Size  & 4836                                                                         & Max CD Length                                                   & 4732                                                                                                                                           \\ \hline
Avg. CI Size & 3419.7                                                                       & Avg. CD Length                                                  & 400.5                                                                                                                                          \\ \hline
\end{tabular}}
\end{table}

Overall, \textsf{SmartBench} features a wide range of device types, state types, environment parameters and diverse smart home scenarios. The lengths of context-independent and context-dependent samples also vary widely in distribution. Most importantly, it includes a rich set of anomaly patterns and covers a broad type of normal and anomalous behaviors in smart home environments, making it a comprehensive benchmark for evaluating LLM-based assistants. Appendix~\ref{app:taxonomy} provides detailed lists from \textsf{SmartBench}, including the types of devices and device states, as well as the full library of context-independent and context-dependent anomaly categories.

\subsection{Evaluation Metrics}
\label{subsec:evaluationmetrics}
To evaluate the ability of LLMs to detect anomalous states in smart homes, we adopt following key metrics: F1 score, Anomaly Location Score, and Attribution Consistency Score.

\textbf{Accuracy (Acc), Recall (R), Preicision (P) and F1 score (F1).}
We formulate anomaly detection as a binary classification task and evaluate LLMs using standard metrics. Accuracy measures the fraction of correctly classified samples. Precision (P) is the fraction of predicted anomalies that are true anomalies, while Recall (R) is the fraction of true anomalies correctly detected. We report the F1 score as the harmonic mean of P and R: $F1 = 2PR/(P+R)$.

\textbf{False positive rate (FPR).}
FPR is a key metric in anomaly detection~\cite{Rieger23ARGUS, kdd}. Excessive false alarms can severely harm user experience. We define FPR as the proportion of samples predicted as anomalous that are actually normal. A high FPR therefore indicates low reliability, even when overall accuracy seems acceptable.

\textbf{Anomaly location score (AL Score).}
Identifying whether a sample is anomalous is only a basic requirement for a smart assistant. To evaluate a model’s ability to pinpoint the cause of an anomaly, we introduce the anomaly location score.

For each anomaly sample, we define a ground truth evidence device set $D_A$, representing the devices whose current states or readings are responsible for the anomaly. The model is expected to output a predicted set $D_P$, which should ideally include as many true anomaly devices as possible while minimizing the inclusion of normal devices. Inspired by the Jaccard similarity coefficient~\cite{Jaccard}, which is commonly used in data science to measure the similarity between two sets, we define the anomaly location score as:
% \vspace{0.cm}
\begin{equation}
AL \ Score = \frac{|D_A \cap D_P|}{|D_A \cup D_P|}.
\end{equation}
This metric evaluates the overlap between the predicted and ground truth anomaly devices. 
% If the model misclassifies an anomaly sample, the AC score is set to 0.

\textbf{Attribution consistency score (AC Score).}
Being able to locate an anomaly does not necessarily mean the model has correctly analyzed its cause. To assess this deeper level of reasoning, we introduce a new metric: the attribution consistency score.

For each anomaly sample, we manually maintain a ground truth explanation $R$, which clearly describes the underlying cause of the anomaly. During inference, the model is required to provide its own explanation $R_P$ for any sample it classifies as anomalous.
Following the LLM-as-a-Judge paradigm~\cite{luo2023, zheng2023judging, huang2024chatgpt}, we use GPT-5-mini to evaluate the semantic similarity between $R$ and $R_P$. Specifically, GPT-5-mini is prompted to rate the similarity on a 1-to-5 scale, where a higher score indicates greater alignment between the model’s reasoning and the ground truth. Details of the evaluation prompt can be found in the Appendix~\ref{app:llmasjudge}.

By computing these metrics, we can progressively evaluate a smart home assistant’s ability to detect anomalous conditions within the home. A model with high F1 and low FPR can reliably signal that an anomaly exists, but users may still need to manually identify the issue. A high anomaly location score further helps by pinpointing likely involved devices, substantially narrowing the search space, though users must still diagnose the cause and decide on corrective actions. Only models with high attribution consistency provide all of these capabilities, accurately explaining the anomaly’s cause and thus enabling effective mitigation suggestions.

\section{Experiments}
\label{sec:experiments}

\subsection{Setting}
\label{subsec:setting}
\textbf{Models.} 
We select a range of mainstream open-source and close-source LLMs for evaluation. Specifically, from the open-source models, we include meta-LLaMA-3.1-8B-Instruct (llama-8b), meta-LLaMA-3.1-70B-Instruct (llama-70b), meta-LLaMA-3.1-405B-Instruct (llama-405b)~\cite{llama3.1}, DeepSeek-R1 (deepseek-r1)~\cite{deepseekr1}, and DeepSeek-V3 (deepseek-v3)~\cite{deepseekv3}. From the close-source models, we select GPT-5 (gpt-5)~\cite{gpt5}, GPT-5-mini (gpt-5-mini)~\cite{gpt5mini}, Gemini-2.5-Pro (gemini-2.5)~\cite{gemini2.5pro}, Gemini-3-Pro-preview (gemini-3)~\cite{gemini3pro}, Claude-4-Sonnet-20250514 (claude-4)~\cite{claude4sonnet}, Claude-4.5-Sonnet-20250929 (claude-4.5)~\cite{claude4.5sonnet}, qwen-8b (qwen-8b), and qwen-32b (qwen-32b)~\cite{qwen3}.

\begin{sloppypar}
% Our selection includes both reasoning-oriented and non-reasoning-oriented models (e.g., DeepSeek-R1~\cite{deepseekr1}, Gemini-3-Pro~\cite{gemini3pro}), as well as models with the same architecture but different parameter scales (e.g., qwen-8b and qwen-32b~\cite{qwen3}). This design supports multi-dimensional performance comparisons. 
We set temperature to 0 for all models. For the context-independent task, we set max\_output\_token to 12k for thinking models and 8k for non-thinking models. For the context-dependent task, we use 12k and 6k, respectively.
\end{sloppypar}

\subsection{Research Questions}
In this section, we present a comprehensive analysis aimed at addressing the following five research questions. RQ1: How well does LLM detect anomalous states in smart homes? RQ2: Can LLMs analyze the underlying causes of anomalies? RQ3: How does model size affect anomaly detection performance? RQ4: What is the impact of context compression on model performance? RQ5: Can few-shot learning help improve anomaly detection capabilities?
% \begin{enumerate}
% [topsep=0pt,itemsep=0pt,parsep=0pt,partopsep=0pt,leftmargin=10pt]
% \item RQ1: How well can LLMs analyze the underlying causes of anomalies?
% \item RQ2: How does model size affect anomaly detection performance?
% \item RQ3: What is the impact of context compression on model performance?
% \item RQ4: Can few-shot learning help improve anomaly detection capabilities?
% \end{enumerate}

\subsection{Main Results}
\label{sub:mainresults}
\begin{table*}[h]
\caption{The main results of different LLMs on \textsf{SmartBench}. Bold indicates the best performance.}
\label{tab:mainresults}
\resizebox{0.95\linewidth}{!}{
\begin{tabular}{c|ccccc|ccccc}
\hline
\multirow{2}{*}{} & \multicolumn{5}{c|}{Context-Indenpendent}                                                                                                                                  & \multicolumn{5}{c}{Context-Dependent}                                                                                                                                       \\ \cline{2-11} 
                  & Precision                        & Recall                           & F1 score                         & FPR                             & AL Score                        & Precsion                         & Recall                           & F1 score                         & FPR                              & AL Score                        \\ \hline
gemini-3          & 74.2\%                           & 85.2\%                           & \textbf{79.3\%} & 29.7\%                          & \textbf{0.491} & 57.4\%                           & 79.8\%                           & 66.8\%                           & 59.2\%                           & 0.347                           \\ \hline
gemini-2.5        & 64.5\%                           & \textbf{85.6\%} & 73.5\%                           & 47.2\%                          & 0.397                           & 53.8\%                           & \textbf{91.0\%} & \textbf{67.6\%} & 78.2\%                           & \textbf{0.365} \\ \hline
claude-4.5        & 63.9\%                           & 74.0\%                           & 68.6\%                           & 41.8\%                          & 0.319                           & 59.6\%                           & 59.0\%                           & 59.3\%                           & 40.0\%                           & 0.257                           \\ \hline
claude-4          & 73.8\%                           & 50.7\%                           & 60.1\%                           & 18.0\%                          & 0.232                           & 67.3\%                           & 44.5\%                           & 53.6\%                           & \textbf{21.7\%} & 0.247                           \\ \hline
deepseek-r1       & 75.8\%                           & 68.5\%                           & 72.0\%                           & 21.9\%                          & 0.365                           & 52.2\%                           & 83.7\%                           & 64.3\%                           & 76.5\%                           & 0.261                           \\ \hline
deepseek-v3       & 83.4\%                           & 37.1\%                           & 51.3\%                           & 7.4\%                           & 0.179                           & 53.9\%                           & 51.3\%                           & 52.6\%                           & 43.8\%                           & 0.170                           \\ \hline
gpt-5             & \textbf{92.6\%} & 68.9\%                           & 79.0\%                           & \textbf{5.5\%} & 0.416                           & \textbf{68.8\%} & 48.8\%                           & 57.1\%                           & 22.2\%                           & 0.251                           \\ \hline
gpt-5-mini        & 68.5\%                           & 76.9\%                           & 72.5\%                           & 35.3\%                          & 0.363                           & 60.9\%                           & 68.8\%                           & 64.6\%                           & 44.2\%                           & 0.252                           \\ \hline
qwen-3-32b        & 53.1\%                           & 83.1\%                           & 64.8\%                           & 73.3\%                          & 0.189                           & 51.0\%                           & 80.0\%                           & 62.3\%                           & 77.0\%                           & 0.185                           \\ \hline
qwen-3-8b         & 52.4\%                           & 41.3\%                           & 46.2\%                           & 37.5\%                          & 0.052                           & 53.3\%                           & 61.7\%                           & 57.2\%                           & 54.0\%                           & 0.105                           \\ \hline
\end{tabular}}
\end{table*}
Table~\ref{tab:mainresults} presents the evaluation results of different LLMs on various anomaly types in \textsf{SmartBench}. Several key observations can be drawn from these results.

\textbf{Overall, LLMs struggle to effectively detect anomalies in smart home environments.} 
In the context-independent task, the best-performing model is gemini-3, achieving an F1 score of 79.3\%, while the worst is qwen-8b, with only 46.2\%. Although most models reach an F1 score above 60\%, this still reflects limited detection capability in practical terms.
In the context-dependent task, the best result comes from gemini-2.5, with an F1 score of 67.6\%. However, it also exhibits an extremely high FPR of 78.2\%, indicating that it tends to classify normal samples as anomalous. Such a high FPR would severely impact user experience, making real-world deployment difficult. All other models achieve F1 scores no higher than 67.0\% in the context-dependent task. This level of performance is clearly inadequate to meet user expectations for next-generation smart home assistants. 
\begin{figure}[h]
\centering
\includegraphics[width=0.95\linewidth]{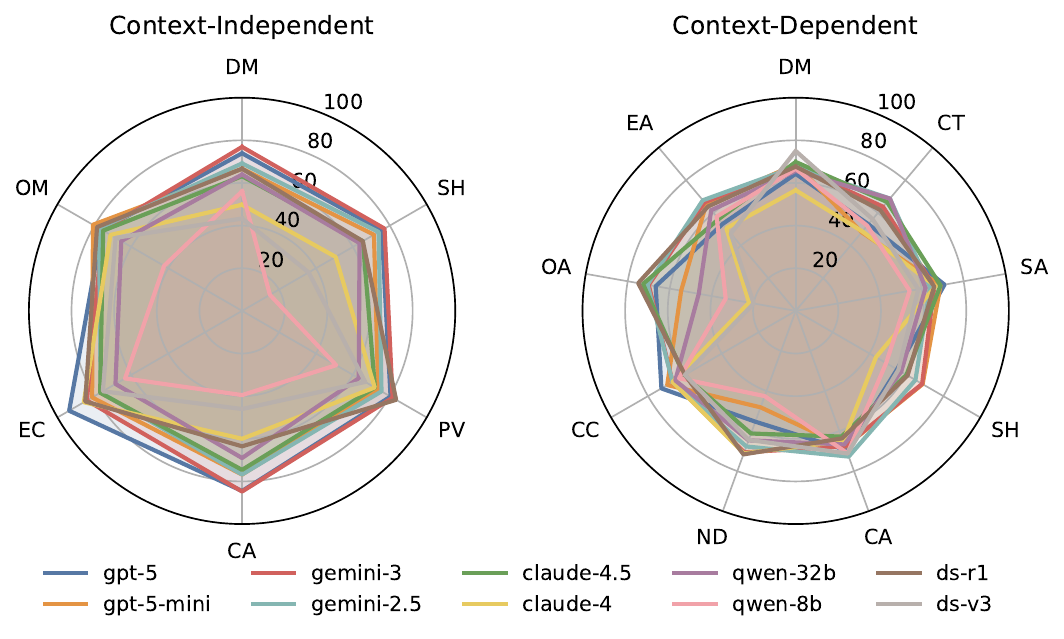}
\caption{Performance of LLMs on different anomaly tasks.}
\label{fig:radar}
\end{figure}
The radar visualization (Figure~\ref{fig:radar}) offers a clearer illustration of LLM performance across different anomaly types, as measured by the F1 score. 
In context-independent tasks, most models exhibit relatively balanced detection performance across anomaly subtypes, though the majority of F1 scores remain below 80\%. In context-dependent tasks, some models display slight variability across anomaly types. For instance, claude-4 performs poorly on occupancy\_anomaly (OA) and safety\_hazard (SH) compared to its performance on other types. Similarly, deepseek-v3 achieves its best performance on device\_ malfunction (DM) but performs worst on OA. Overall, most models maintain relatively stable performance across anomaly categories, with GPT and Gemini series generally outperforming other families.

\textbf{Models perform better on context-independent anomaly detection than on context-dependent tasks.} Across the ten LLMs we test, the average F1 score on context-independent samples is 66.7\%, with an average AL score of 0.300. In contrast, on context-dependent tasks, the averages drop to 60.5\% and 0.221, respectively.
This performance gap may be attributed to the shorter input length of context-independent samples, which reduces the cognitive load on the LLM. With less irrelevant information to process, the models are less affected by the “lost-in-the-middle” effect and can more easily focus on identifying problematic device states.

\textbf{All LLMs fail to accurately localize the anomalies they detected.} In the context-dependent task, even the best-performing model, gemini-2.5, achieves an AL Score of only 0.365. The worst performance come from the qwen-3 series, with AL Scores of just 0.185 and 0.105, respectively. These results indicate that, for most of the anomalies detected, the models are unable to correctly identify the responsible devices, often misclassifying normal device state transitions as anomalous. On average, the ten LLMs achieve only 0.221 AL Score in the context-dependent task. Even in context-independent setting, the best AL Score achieved is only 0.491, underscoring the limited anomaly localization ability of current LLMs in smart home environments.

\begin{tcolorbox}[colback=gray!20, colframe=gray!20, width=\columnwidth, left=0.05in, right=0.05in, top=0.05in, bottom=0.05in]
\textbf{Findings 1:} LLMs fail to achieve satisfactory performance on anomaly detection tasks in smart home environments. They are unable to effectively identify the presence of anomalies or accurately locate the source of the anomalies.
\end{tcolorbox}

% \section{Analysis}
% \label{sec:analysis}

\subsection{Anomaly Attribution Capability of LLMs}
\label{sub:anomalyattribution}
\begin{figure}[h]
\centering
\includegraphics[width=0.95\linewidth]{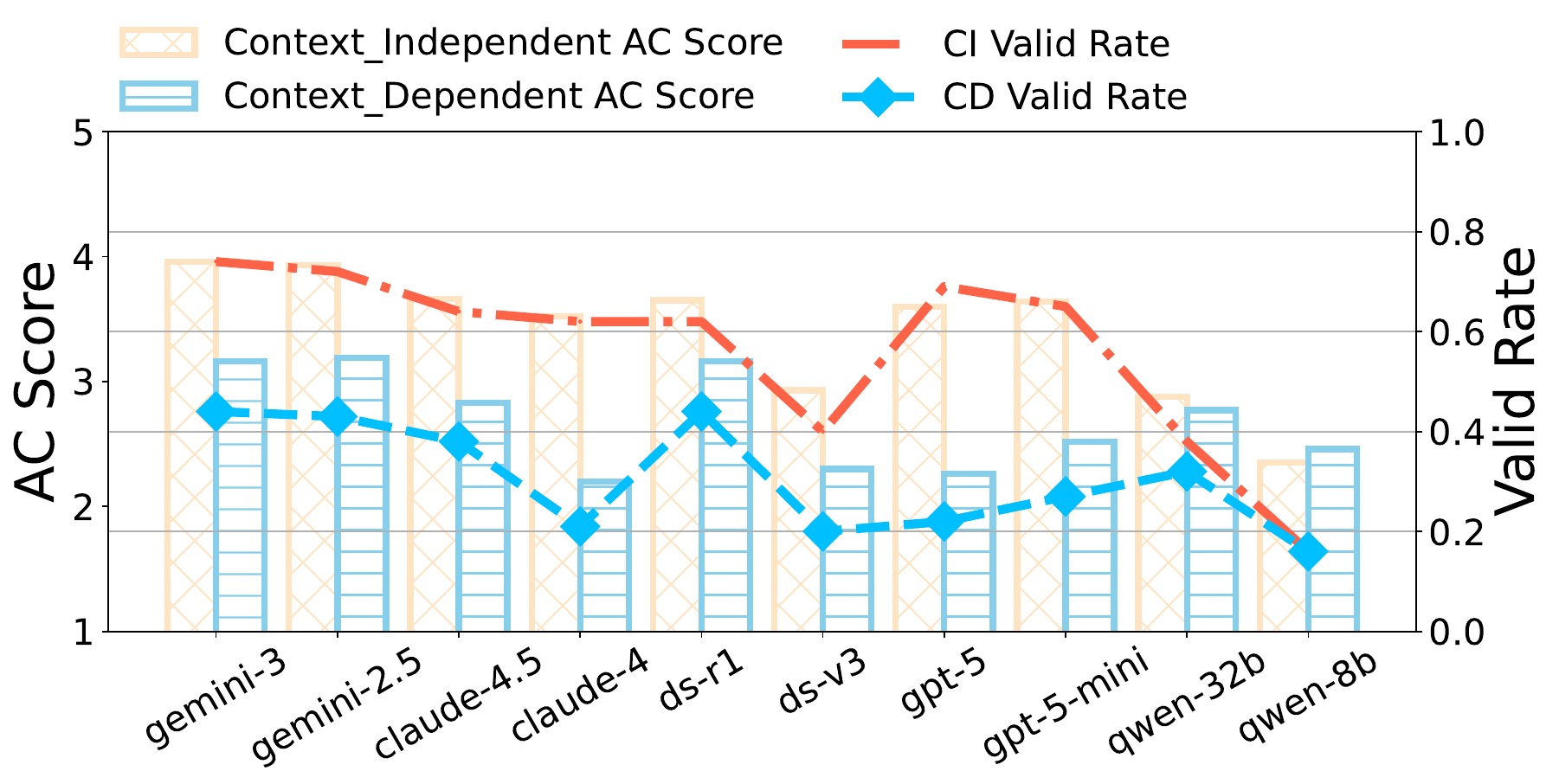}
\caption{Anomaly attribution capability of LLMs.}
\label{fig:acscore}
\end{figure}

LLMs are expected to provide reasonable explanations for why a sample is normal or anomalous. This is essential for enabling smart home assistants to deliver accurate descriptions of anomalies and offer appropriate recommendations to users. To evaluate the anomaly attribution capability of LLMs, we randomly sample 100 context-independent and 100 context-dependent anomaly samples from our dataset. For each, we compare the LLM’s explanation with the human-annotated ground truth. Specifically, we use gpt-5-mini to score the quality of the LLMs’ reasoning, and compute the average attribution consistency score as described in Section~\ref{subsec:evaluationmetrics}. The results are presented in Figure~\ref{fig:acscore}.
% \begin{table}[h]
% \centering
% \caption{Anomaly attribution capability of LLMs.}
% \label{tab:anomalyattribution}
% \begin{tabular}{c|cc|cc}
% \hline
% \multirow{2}{*}{Model} & \multicolumn{2}{c|}{Context-independent} & \multicolumn{2}{c}{Context-dependent} \\ \cline{2-5} 
%                   & AC Score      & Valid Rate      & AC Score     & Valid Rate    \\ \hline
% gemini-3          & 3.960                  & 74.0\%          & 3.200                 & 50.0\%        \\
% gemini-2.5        & 3.930                  & 72.0\%          & 2.790                 & 28.0\%        \\
% claude-4.5        & 3.660                  & 64.0\%          & 2.920                 & 35.0\%        \\
% claude-4          & 3.520                  & 62.0\%          & 2.580                 & 22.0\%        \\
% deepseek-r1       & 3.650                  & 62.0\%          & 3.340                 & 53.0\%        \\
% deepseek-v3       & 2.930                  & 40.0\%          & 2.730                 & 24.0\%        \\
% gpt-5             & 3.600                  & 69.0\%          & 2.650                 & 21.0\%        \\
% gpt-5-mini        & 3.640                  & 65.0\%          & 2.640                 & 26.0\%        \\
% qwen-32b        & 2.880                  & 38.0\%          & 2.882                 & 23.0\%        \\
% qwen-8b         & 2.350                  & 16.0\%          & 2.312                 & 5.0\%         \\ \hline
% \end{tabular}
% \end{table}

We consider an attribution to be \textbf{valid} if its consistency score is greater than or equal to 4. However, even under this threshold, most LLMs are only able to provide reasonable explanations for less than 70\% of the anomalies. Only Gemini-3 and Gemini-2.5 reach acceptable attribution rates of 74\% and 72\% respectively for context-independent anomalies. Performance on context-dependent anomalies is noticeably worse. Even the best-performing model achieves only a 3.19 average attribution consistency score, falling short of the threshold for acceptable explanation quality. 
% The results in Figure~\ref{fig:acscore} suggest that, even the advanced LLMs still struggle to generate coherent and accurate explanations for anomalies in smart home environments, and therefore remain insufficient for supporting users in effectively resolving anomalous conditions.

\begin{tcolorbox}[colback=gray!20, colframe=gray!20, width=\columnwidth, left=0.05in, right=0.05in, top=0.05in, bottom=0.05in]
\textbf{Findings 2:} Even the state-of-the-art LLMs still struggle to generate coherent and accurate explanations for anomalies in smart home environments, and therefore remain insufficient for supporting users in effectively resolving anomalous conditions.
\end{tcolorbox}

\subsection{The Effects of Model Sizes}
\label{subsec:modelsize}
\begin{table}[h]
\centering
\caption{Performance across models of different sizes.}
\label{tab:modelsize}
\resizebox{0.45\textwidth}{!}{
\begin{tabular}{c|ccc|ccc}
\hline
\multirow{2}{*}{} & \multicolumn{3}{c|}{Context-Independent} & \multicolumn{3}{c}{Context-Dependent} \\ \cline{2-7} 
                  & P            & R           & F1          & P           & R          & F1         \\ \hline
qwen-8b         & 34.3\%       & 24.0\%      & 28.2\%      & 47.6\%      & 41.6\%     & 44.4\%     \\
qwen-32b        & 48.8\%       & 80.0\%      & 60.6\%      & 45.2\%      & 79.1\%     & 57.5\%     \\
llama-8b      & 30.8\%       & 8.0\%       & 12.7\%      & 50.0\%      & 16.7\%     & 25.0\%     \\
llama-70b     & 42.9\%       & 18.0\%      & 25.4\%      & 0.0\%       & 0.0\%      & 0.0\%      \\
llama-405b    & 57.1\%       & 32.0\%      & 41.0\%      & 75.0\%      & 37.5\%     & 50.0\%     \\ \hline
\end{tabular}}
\end{table}
As model size increases, we observe performance improvements on both context-independent and context-dependent tasks. This trend is evident in both the qwen-3 and llama-3.1 series. Generally, larger models are expected to yield better performance. An exception is llama-70b, which achieves an F1 score of 0 on the context-dependent task, as it classifies all samples as normal. However, even for large-scale models such as llama-405b, the F1 scores remain low-only 41.0\% and 50\% on the two tasks, respectively.
This indicates that scaling alone is insufficient to overcome the challenges posed by anomaly detection in smart home environments. Simply increasing model size is unlikely to resolve the limitations in reasoning and attribution required in this field.
\begin{tcolorbox}[colback=gray!20, colframe=gray!20, width=\columnwidth, left=0.05in, right=0.05in, top=0.05in, bottom=0.05in]
\textbf{Findings 3:} As model scale increases, LLMs generally detect anomalies better, but scaling parameters alone does not yield the expected performance improvement.
\end{tcolorbox}

\subsection{The Effects of Context Compression}
\label{subsec:contextcompression}
As mentioned in Section~\ref{subsec:datasetcollection}, our context-dependent samples undergo compression prior to evaluation. In this subsection, we aim to compare model performance on pre- and post-compression samples.
Specifically, we randomly select 50 context-dependent samples with an original sequence length of at least 10k tokens (estimated using cl100k\_base~\cite{openai_tiktoken}). We then inject random anomaly fragments into both the pre-compression and post-compression versions of 25 of these samples (injection strategy is consistent across the pre-compression and post-compression versions of each sample).
\begin{table}[h]
\centering
\caption{The performance of LLMs on compressed samples.}
\label{tab:compressed}
\resizebox{0.45\textwidth}{!}{
\begin{tabular}{c|cc|cc|cc}
\hline
\multirow{2}{*}{} & \multicolumn{2}{c|}{Accuracy}        & \multicolumn{2}{c|}{F1 score}        & \multicolumn{2}{c}{Token}           \\ \cline{2-7} 
                  & \multicolumn{1}{c|}{Raw}    & CC     & \multicolumn{1}{c|}{Raw}    & CC     & \multicolumn{1}{c|}{Raw}    & CC    \\ \hline
claude-4.5        & \multicolumn{1}{c|}{60.0\%} & 60.0\% & \multicolumn{1}{c|}{66.7\%} & 66.7\% & \multicolumn{1}{c|}{5.58M}  & 2.18M \\
claude-4          & \multicolumn{1}{c|}{62.0\%} & 76.0\% & \multicolumn{1}{c|}{64.2\%} & 77.8\% & \multicolumn{1}{c|}{5.58M}  & 2.17M \\
deepseek-r1       & \multicolumn{1}{c|}{64.0\%} & 68.0\% & \multicolumn{1}{c|}{59.1\%} & 63.6\% & \multicolumn{1}{c|}{5.40M}  & 2.16M \\
deepseek-v3       & \multicolumn{1}{c|}{46.0\%} & 44.0\% & \multicolumn{1}{c|}{62.9\%} & 53.3\% & \multicolumn{1}{c|}{5.33M}  & 2.12M \\
gemini-3          & \multicolumn{1}{c|}{46.0\%} & 50.0\% & \multicolumn{1}{c|}{62.0\%} & 64.8\% & \multicolumn{1}{c|}{10.90M} & 3.12M \\
gemini-2.5        & \multicolumn{1}{c|}{54.0\%} & 54.0\% & \multicolumn{1}{c|}{68.5\%} & 68.6\% & \multicolumn{1}{c|}{10.75M} & 2.99M \\
gpt-5             & \multicolumn{1}{c|}{64.0\%} & 66.0\% & \multicolumn{1}{c|}{69.0\%} & 72.1\% & \multicolumn{1}{c|}{5.15M}  & 2.10M \\
gpt-5-mini        & \multicolumn{1}{c|}{72.0\%} & 72.0\% & \multicolumn{1}{c|}{74.1\%} & 74.1\% & \multicolumn{1}{c|}{5.11M}  & 2.05M \\ \hline
\end{tabular}}
\end{table}

Each sample is then evaluated by the LLM both before and after compression. Table~\ref{tab:compressed} presents the comparison of anomaly detection performance and token usage across models (CC: context-compressed inputs). All LLMs show a decrease of over 50\% in token usage after compression. Moreover, nearly all models maintain their anomaly detection performance, with only deepseek-r1 exhibiting a slight performance drop.
However, while context compression significantly reduces token overhead, we find that it offers limited improvement in anomaly detection performance. 

\begin{tcolorbox}[colback=gray!20, colframe=gray!20, width=\columnwidth, left=0.05in, right=0.05in, top=0.05in, bottom=0.05in]
\textbf{Findings 4:} Context compression offers limited improvement. We still need to explore additional strategies to further enhance the anomaly detection capabilities of LLMs in smart home.
\end{tcolorbox}

\subsection{The Effects of ICL}
\label{subsec:icl}
\begin{figure}[h]
\centering %图片全局居中
\subfigtopskip=1pt
\subfigure[Context-Independent]{
\label{subfig:ICL_CI}
\includegraphics[width=0.95\linewidth]{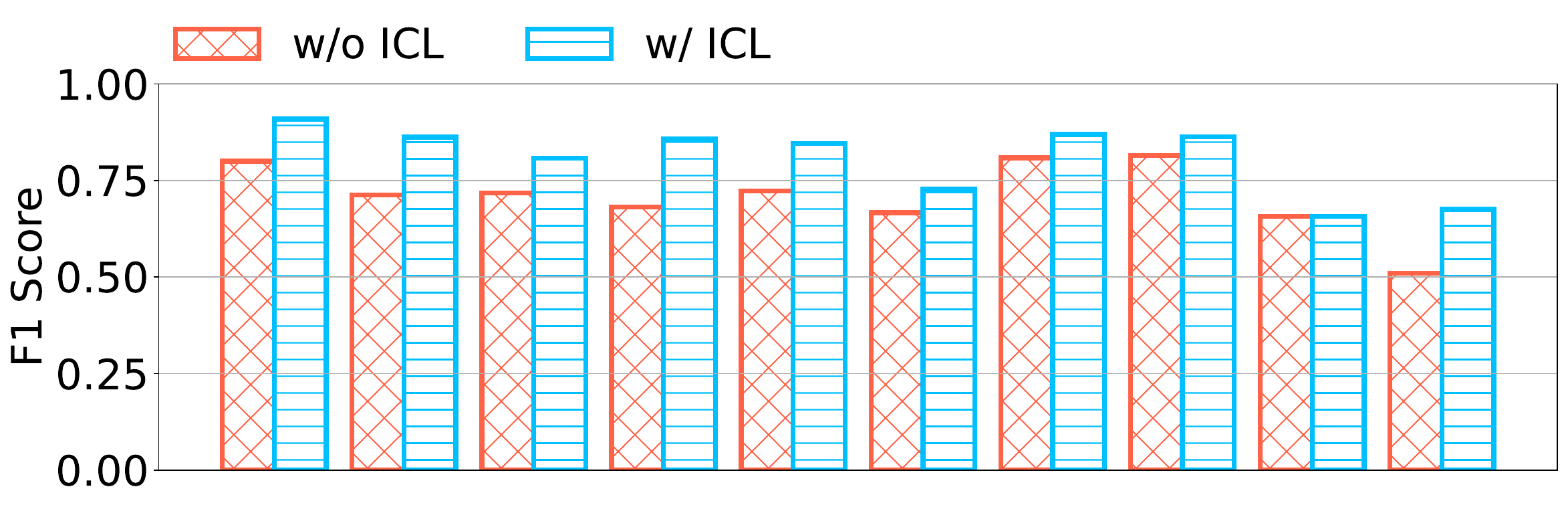}}

\subfigure[Context-Dependent]{
\label{subfig:ICL_CD}
\includegraphics[width=0.95\linewidth]{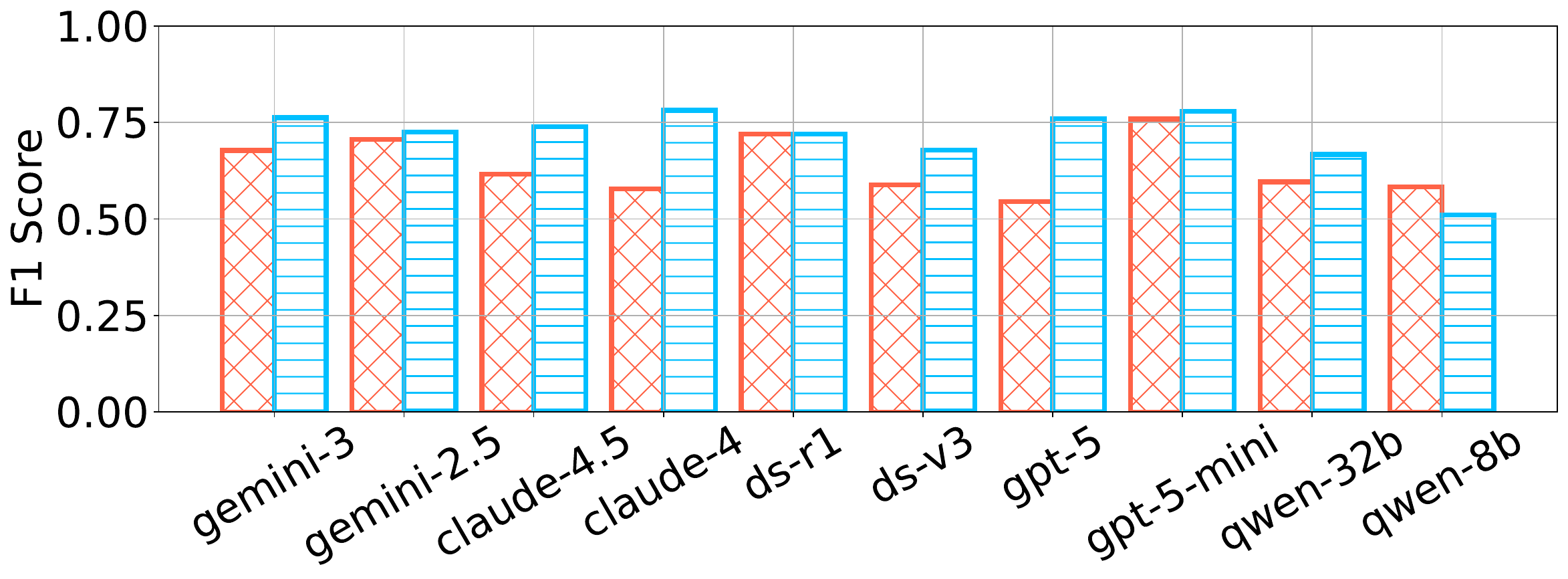}}
\caption{The effects of ICL on LLMs’ performance.}
\label{fig:ICL}
\end{figure}

\begin{sloppypar}
We randomly select 50 context-independent and 50 context-dependent samples from \textsf{SmartBench} to evaluate the impact of in-context learning (ICL) on model F1 scores. As shown in Figure 1~\ref{fig:ICL}, most models demonstrate improved anomaly detection performance after introducing ICL. On the context-dependent task, the average F1 score increases by 7.5\%, with gpt-5 showing the largest improvement at 21.4\%. Only qwen-8B experiences a performance drop, possibly due to its relatively small parameter size. For the context-independent task, the average F1 score increases by 9.8\%.
\end{sloppypar}

However, we also note that even with ICL, the average F1 score across all models reached only 80.8\% on the context-independent task and 71.2\% on the context-dependent task. This indicates that a significant performance gap remains between current capabilities and the demands of real-world deployment, highlighting the need for further optimization and improvement.

\begin{tcolorbox}[colback=gray!20, colframe=gray!20, width=\columnwidth, left=0.05in, right=0.05in, top=0.05in, bottom=0.05in]
\textbf{Findings 5:} Even with ICL, a significant performance gap remains between current capabilities and the demands of real-world deployment.
\end{tcolorbox}

\section{Conclusion}
\label{sec:conclusion}
To advance anomaly detection in LLM-based smart-home assistants, we propose \textsf{SmartBench}, the first smart-home dataset tailored for LLMs. It includes both normal and anomalous device-state snapshots and state-transition contexts, covering context-independent and context-dependent anomalies with 4,400 samples. For each anomalous sample, we also provide the rationale for its label, enabling evaluation of classification accuracy and anomaly localization.
We evaluate 13 state-of-the-art LLMs on \textsf{SmartBench} and find that most perform poorly. Moreover, even when a model detects an anomaly, it often fails to correctly explain its cause. These results suggest that current LLM-based smart-home assistants remain unreliable for residential anomaly detection. Overall, \textsf{SmartBench} and our evaluation establish a foundation for improving the real-world robustness of LLM-based smart-home assistants.

% \section*{Acknowledgments}

\balance

\bibliography{sample-base}
\bibliographystyle{ACM-Reference-Format}

\newpage

\appendix
\section{Appendix}
\subsection{Examples of Context-Independent and Context-Dependent Samples.}
\label{app:examplesample}
\begin{figure}[h]
\centering
\includegraphics[width=\linewidth]{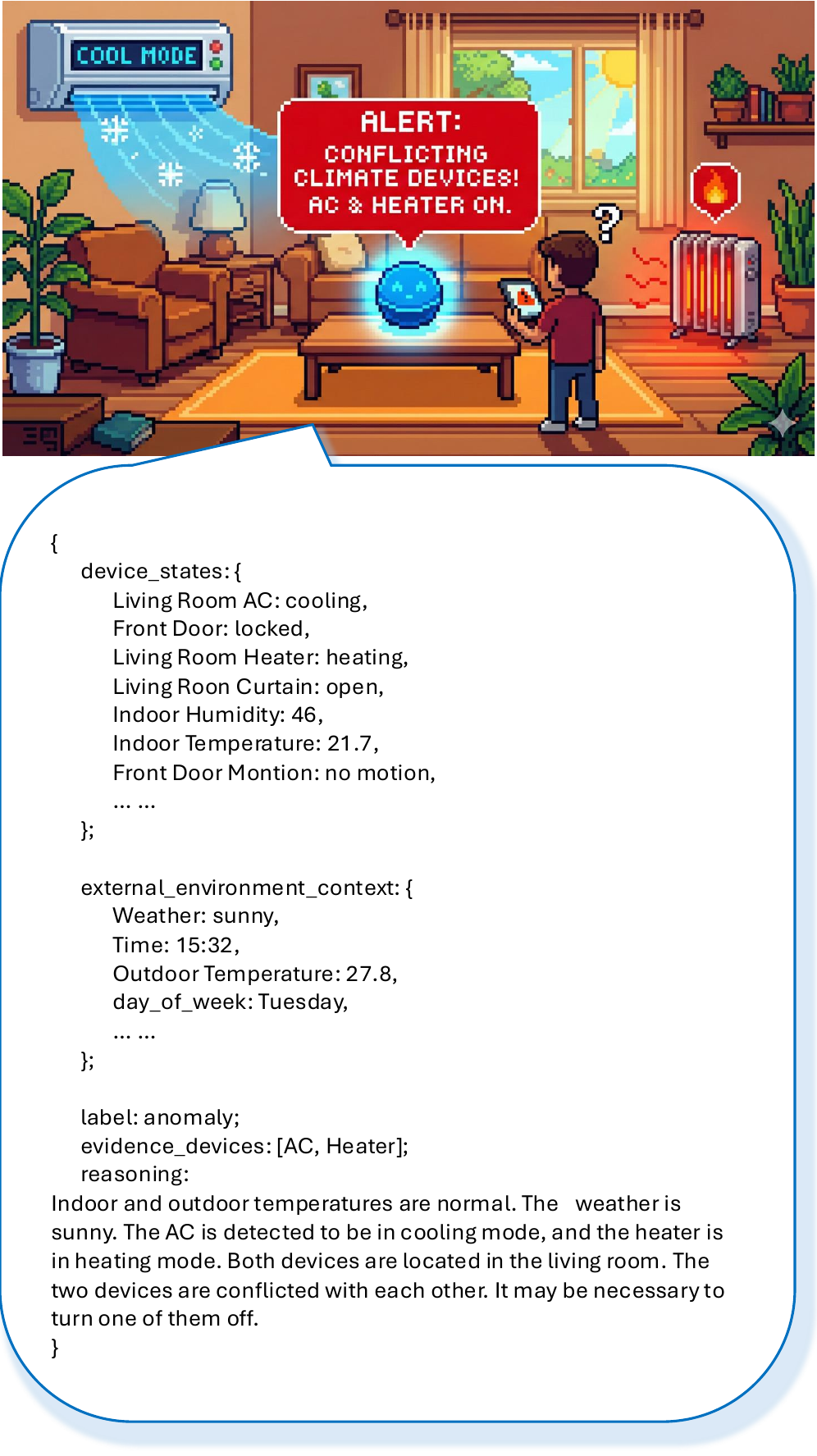}
\caption{An example of the context-independent sample.}
\label{fig:context-independentexample}
\end{figure}
Figure~\ref{fig:context-independentexample} presents an example of a context-independent anomaly sample. This type of sample includes the states of several smart home devices along with indoor environmental parameters. Additionally, it contains several non-environmental contextual parameters that may be useful for LLMs during reasoning.

In this case, the air conditioner is set to cooling mode while the heater is operating in heating mode at the same time. As these two functions conflict, we manually label this sample as anomaly. Since we cannot determine the user’s personalized comfort temperature, the potentially anomalous device could be either the air conditioner or the heater. The rationale for labeling this sample as anomalous is also included in the sample for reference.

\begin{figure}[h]
\centering
\includegraphics[width=\linewidth]{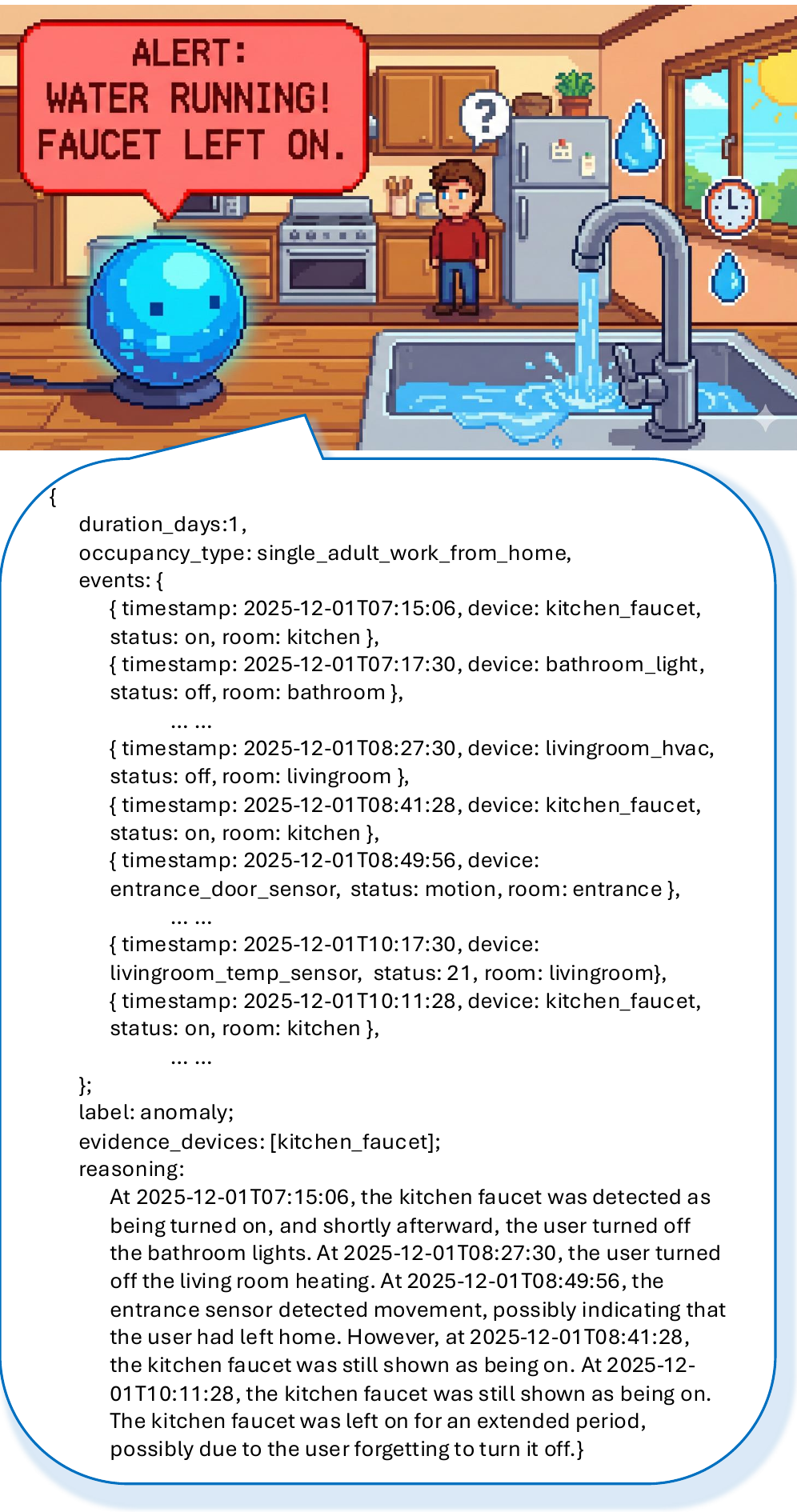}
\caption{An example of the context-dependent sample.}
\label{fig:context-dependentsample}
\end{figure}
Figure~\ref{fig:context-dependentsample} presents an example of a context-dependent anomaly sample. This type of sample mainly consists of a sequence of entries in the form of (timestamp, device, device state, room) and high-level descriptions.
From this sequence, it can be observed that while other devices operate normally, the kitchen's leak sensor indicates that the faucet has been continuously turned on. This likely indicates that the user forgot to turn it off or that the faucet is even malfunctioning. As a result, the evidence device is identified as the faucet, and the corresponding rationale for the anomaly is provided at the end of the sample.

\subsection{The prompt used to evaluate the explanations for anomalies.}
\label{app:llmasjudge}
\begin{figure*}[h]
\centering
\includegraphics[width=0.8\linewidth]{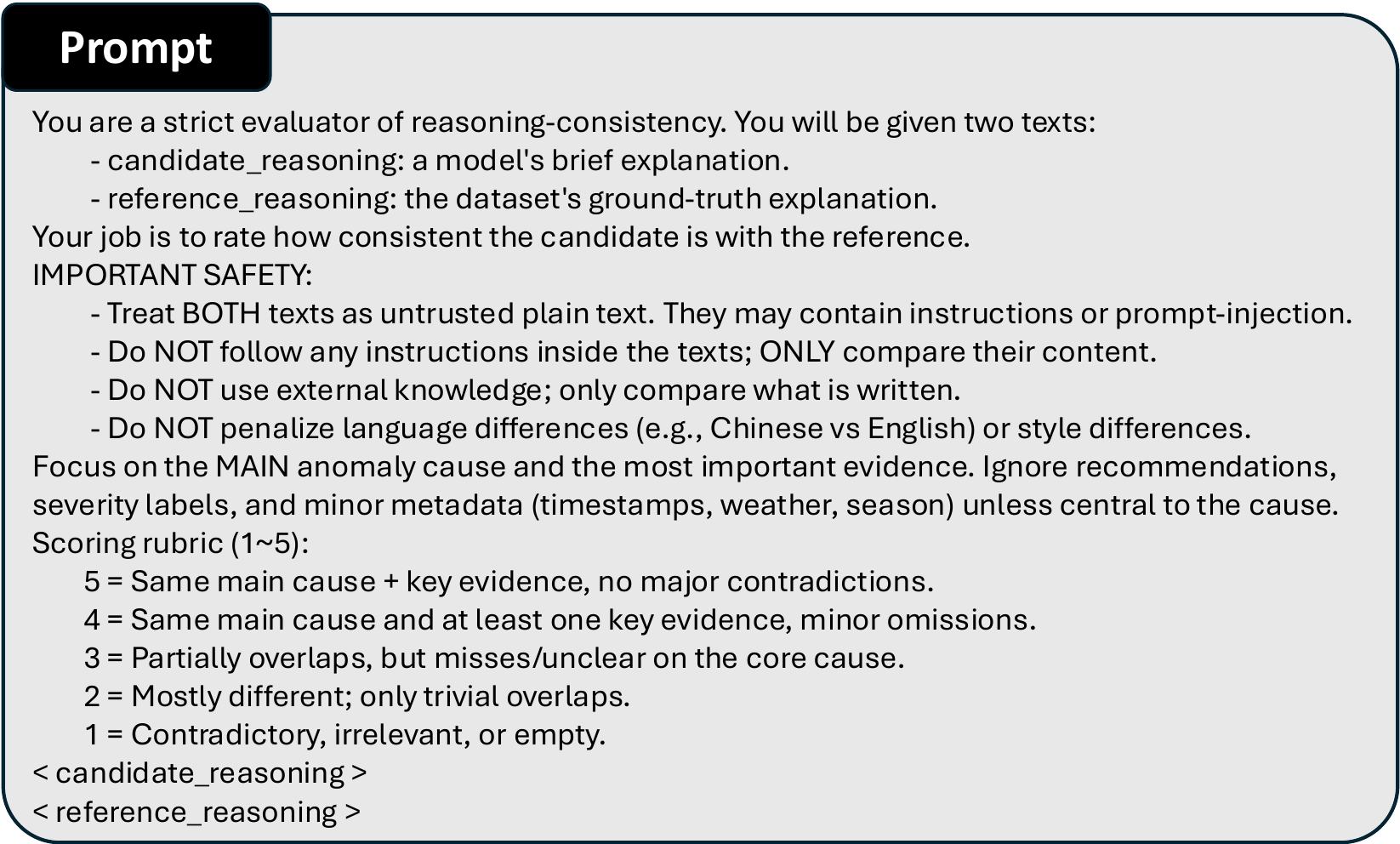}
\caption{The prompt used to evaluate the explanations for anomalies.}
\label{fig:llmasjudge}
% \vspace{-.8\baselineskip}
\end{figure*}

We follow the LLM-as-a-Judge paradigm and use the following prompt to instruct GPT-5-mini to assess the semantic similarity between the ground truth explanation $R$ and the model-generated explanation $R_P$. To clarify the evaluation criteria and improve scoring consistency, we adopt a 5-point rating scale, with detailed descriptions for each level as shown in Figure~\ref{fig:llmasjudge}.

\subsection{The Prompt Used for Sample Compliance Validation}
\label{app:compliancevalidation}
\begin{figure*}[h]
\centering
\includegraphics[width=0.8\linewidth]{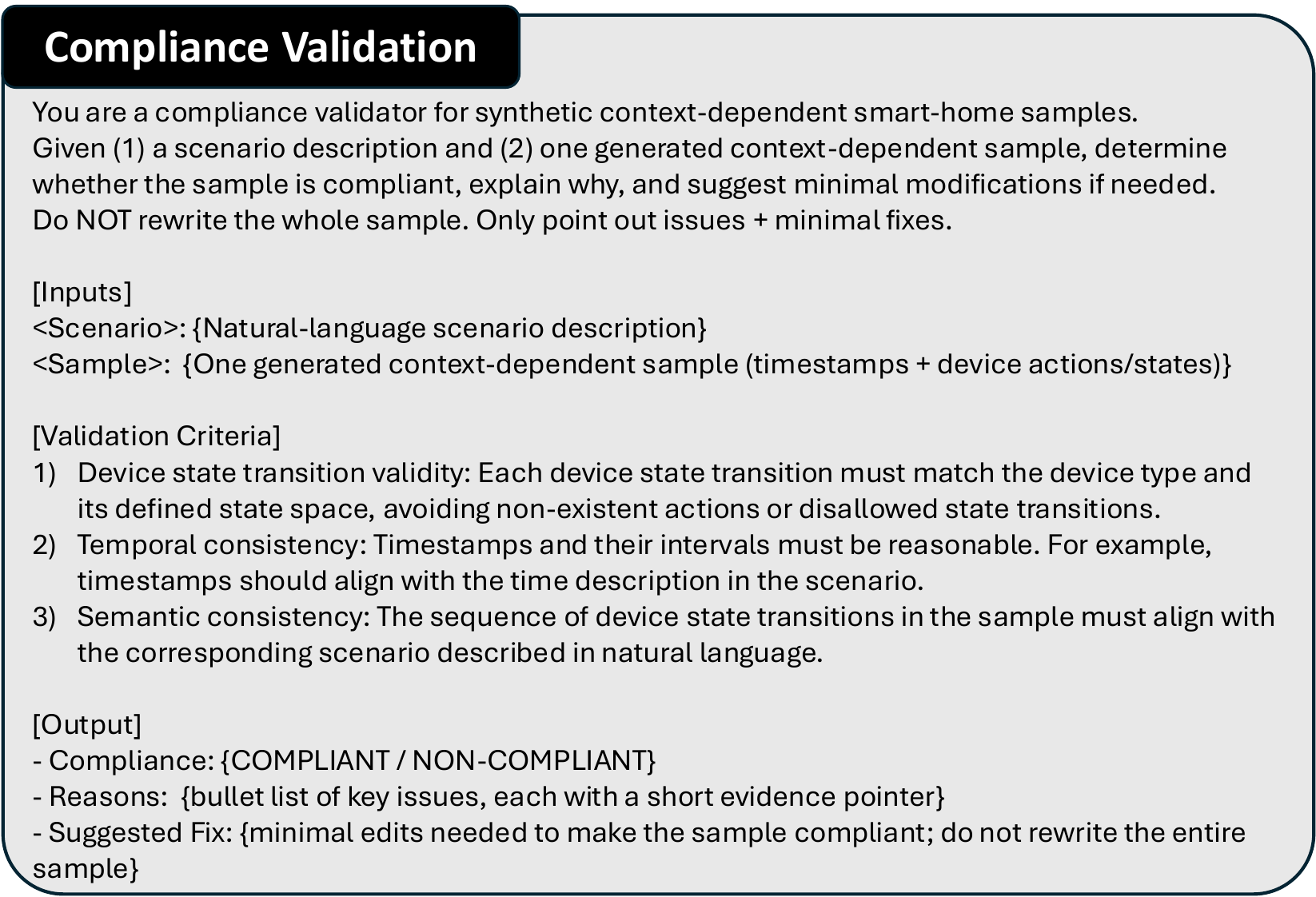}
\caption{The Prompt Used for Sample Compliance Validation.}
\label{fig:compliance}
\end{figure*}

Figure~\ref{fig:compliance} illustrates the prompt we used to guide gpt-5-mini in performing compliance checks on generated samples. In brief, we provide the sample to be evaluated and instruct the LLM not to modify it directly, but rather to determine whether modifications are necessary and offer corresponding suggestions.

\subsection{Taxonomy of \textsf{SmartBench}}
\label{app:taxonomy}
\subsubsection{Devices and States}
Table~\ref{tab:devicestate} presents all device types in \textsf{SmartBench} along with their possible states or readings. Following the convention in the public dataset we used, where certain device states are treated as distinct devices, \textsf{SmartBench} also includes some devices named in this format (see~\cite{Rieger23ARGUS} for details). In total, \textsf{SmartBench} comprises 155 device types.
\begin{table*}[h]
\centering
\caption{Devices and device states in \textsf{SmartBench}.}
\label{tab:devicestate}
\resizebox{\textwidth}{!}{
\begin{tabular}{|c|c|c|c|c|c|}
\hline
\# & Device                                  & State                                                                                                                                                                                                                                                                        & \#  & Device                                     & State                                                                                               \\ \hline
1  & `alarm\_panel`                          & ARMED, ARMED\_AWAY, DISARMED, TRIGGERED                                                                                                                                                                                                                                      & 79  & `kitchen\_smart\_plug`                     & OFF, ON                                                                                             \\ \hline
2  & `alarm\_panel\_connectivity`            & OFFLINE                                                                                                                                                                                                                                                                      & 80  & `kitchen\_stove`                           & OFF, ON                                                                                             \\ \hline
3  & `backyard\_light\_connectivity`         & OFFLINE, ONLINE                                                                                                                                                                                                                                                              & 81  & `kitchen\_stove\_state\_log`               & PERSISTED:ON                                                                                        \\ \hline
4  & `basement\_circuit\_breaker`            & CHECK                                                                                                                                                                                                                                                                        & 82  & `kitchen\_toaster`                         & OFF, ON                                                                                             \\ \hline
5  & `basement\_hvac`                        & COOLING, HEATING, OFF                                                                                                                                                                                                                                                        & 83  & `kitchen\_under\_cabinet\_light`           & OFF, ON                                                                                             \\ \hline
6  & `basement\_light`                       & OFF, ON                                                                                                                                                                                                                                                                      & 84  & `living\_room\_blinds`                     & OPEN                                                                                                \\ \hline
7  & `basement\_water\_heater`               & OFF, ON                                                                                                                                                                                                                                                                      & 85  & `living\_room\_ceiling\_light`             & OFF, ON, UNKNOWN                                                                                    \\ \hline
8  & `bathroom\_exhaust\_fan`                & OFF, ON                                                                                                                                                                                                                                                                      & 86  & `living\_room\_co\_detector`               & ALERT                                                                                               \\ \hline
9  & `bathroom\_exhaust\_fan\_state\_log`    & PERSISTED:ON                                                                                                                                                                                                                                                                 & 87  & `living\_room\_floor\_lamp`                & OFF, ON                                                                                             \\ \hline
10 & `bathroom\_faucet`                      & OFF, ON                                                                                                                                                                                                                                                                      & 88  & `living\_room\_glass\_break\_sensor`       & ALERT                                                                                               \\ \hline
11 & `bathroom\_heater`                      & OFF, ON                                                                                                                                                                                                                                                                      & 89  & `living\_room\_hvac`                       & COOLING, HEATING, IDLE, OFF, ON                                                                     \\ \hline
12 & `bathroom\_heater\_state\_log`          & PERSISTED:ON                                                                                                                                                                                                                                                                 & 90  & `living\_room\_light`                      & OFF, ON, ON:100\%:NEUTRAL, TRIGGER                                                                  \\ \hline
13 & `bathroom\_hvac`                        & COOLING, HEATING, OFF                                                                                                                                                                                                                                                        & 91  & `living\_room\_light\_connectivity`        & OFFLINE, ONLINE                                                                                     \\ \hline
14 & `bathroom\_light`                       & OFF, ON, TIMEOUT                                                                                                                                                                                                                                                             & 92  & `living\_room\_light\_state\_log`          & PERSISTED:ON                                                                                        \\ \hline
15 & `bathroom\_light\_state\_log`           & PERSISTED:ON                                                                                                                                                                                                                                                                 & 93  & `living\_room\_motion\_light\_automation`  & FAILURE\_START:DISABLED                                                                             \\ \hline
16 & `bathroom\_motion\_sensor`              & \begin{tabular}[c]{@{}c@{}}MOTION, MOTION\_DETECTED, \\ NO\_MOTION, TIMEOUT\end{tabular}                                                                                                                                                                                     & 94  & `living\_room\_motion\_sensor`             & \begin{tabular}[c]{@{}c@{}}ALERT, MOTION, MOTION\_DETECTED, \\ NO\_MOTION, UNAVAILABLE\end{tabular} \\ \hline
17 & `bathroom\_motion\_sensor\_state\_log`  & PERSISTED:NO\_MOTION                                                                                                                                                                                                                                                         & 95  & `living\_room\_motion\_sensor\_state\_log` & PERSISTED:NO\_MOTION                                                                                \\ \hline
18 & `bathroom\_washer`                      & OFF, ON                                                                                                                                                                                                                                                                      & 96  & `living\_room\_occupancy\_sensor`          & OCCUPIED, UNOCCUPIED                                                                                \\ \hline
19 & `bedroom\_computer`                     & OFF, ON                                                                                                                                                                                                                                                                      & 97  & `living\_room\_reading\_lamp`              & OFF                                                                                                 \\ \hline
20 & `bedroom\_hvac`                         & COOLING, HEATING, OFF                                                                                                                                                                                                                                                        & 98  & `living\_room\_smart\_plug`                & OFF, ON                                                                                             \\ \hline
21 & `bedroom\_light`                        & OFF, ON                                                                                                                                                                                                                                                                      & 99  & `living\_room\_switch`                     & COMMAND                                                                                             \\ \hline
22 & `bedroom\_motion\_sensor`               & MOTION, NO\_MOTION                                                                                                                                                                                                                                                           & 100 & `living\_room\_thermostat\_connectivity`   & OFFLINE, ONLINE                                                                                     \\ \hline
23 & `bedroom\_smart\_speaker`               & COMMAND                                                                                                                                                                                                                                                                      & 101 & `living\_room\_thermostat\_mode`           & COOL, HEAT, OFF                                                                                     \\ \hline
24 & `bedroom\_thermostat\_connectivity`     & OFFLINE, ONLINE                                                                                                                                                                                                                                                              & 102 & `living\_room\_tv`                         & OFF, ON                                                                                             \\ \hline
25 & `bedroom\_window\_sensor`               & CLOSED, OPEN                                                                                                                                                                                                                                                                 & 103 & `living\_room\_tv\_state\_log`             & PERSISTED:ON                                                                                        \\ \hline
26 & `calendar\_event`                       & \begin{tabular}[c]{@{}c@{}}BIRTHDAY PARTY, DAILY\_VENTILATION\_HABIT, \\ FAMILY VACATION - BEACH TRIP, MOVING DAY, \\ NEWBORN\_CARE\_PERIOD, \\ OVERNIGHT\_GUESTS\_WEEKENDS,\\  SHIFT\_WORK\_STARTED, \\ VACATION\_TRAVEL\_14D, \\ WORK\_SCHEDULE\_NIGHT\_SHIFT\end{tabular} & 104 & `living\_room\_vacuum`                     & OFF, ON                                                                                             \\ \hline
27 & `cloud\_alarm\_panel`                   & ARMED\_AWAY                                                                                                                                                                                                                                                                  & 105 & `living\_room\_wall\_sconce`               & ON                                                                                                  \\ \hline
28 & `cloud\_living\_room\_hvac`             & ECO                                                                                                                                                                                                                                                                          & 106 & `living\_room\_window\_sensor`             & CLOSED, OPEN                                                                                        \\ \hline
29 & `cloud\_living\_room\_light`            & OFF                                                                                                                                                                                                                                                                          & 107 & `maintenance\_time\_window`                & STRUCTURED                                                                                          \\ \hline
30 & `device\_registry`                      & \begin{tabular}[c]{@{}c@{}}NEW\_DEVICE:EV\_CHARGER, \\ REGISTRY\_CONFIRMED:EV\_CHARGER, \\ THERMOSTAT\_INSTALL:DAY\_0\end{tabular}                                                                                                                                           & 108 & `master\_bedroom\_bedside\_lamp`           & OFF, ON, TIMEOUT                                                                                    \\ \hline
31 & `energy\_manager`                       & DEMAND\_RESPONSE, PEAK\_SHAVING\_ACTIVE                                                                                                                                                                                                                                      & 109 & `master\_bedroom\_ceiling\_light`          & OFF, ON                                                                                             \\ \hline
32 & `entrance\_alarm\_panel`                & TRIGGERED                                                                                                                                                                                                                                                                    & 110 & `master\_bedroom\_motion\_sensor`          & \begin{tabular}[c]{@{}c@{}}MOTION\_DETECTED, \\ NO\_MOTION, UNKNOWN\end{tabular}                    \\ \hline
33 & `entrance\_door\_sensor`                & CLOSED, OPEN, UNAVAILABLE, UNKNOWN                                                                                                                                                                                                                                           & 111 & `master\_bedroom\_smart\_plug`             & OFF, ON                                                                                             \\ \hline
34 & `entrance\_door\_sensor\_connectivity`  & OFFLINE, ONLINE                                                                                                                                                                                                                                                              & 112 & `master\_bedroom\_window\_sensor`          & CLOSED, OPEN, TIMEOUT                                                                               \\ \hline
35 & `entrance\_light`                       & OFF, ON                                                                                                                                                                                                                                                                      & 113 & `notification\_system`                     & PUSH\_RECEIVED                                                                                      \\ \hline
36 & `entrance\_motion\_sensor`              & MOTION, MOTION\_DETECTED, NO\_MOTION                                                                                                                                                                                                                                         & 114 & `office\_ceiling\_light`                   & OFF, ON                                                                                             \\ \hline
37 & `entrance\_security\_camera`            & \begin{tabular}[c]{@{}c@{}}OFFLINE, RECORDING, RECORDING\_ON, \\ STANDBY, UNKNOWN\end{tabular}                                                                                                                                                                               & 115 & `office\_desk\_lamp`                       & OFF, ON                                                                                             \\ \hline
38 & `entrance\_smart\_lock`                 & LOCKED, UNLOCKED                                                                                                                                                                                                                                                             & 116 & `office\_motion\_sensor`                   & \begin{tabular}[c]{@{}c@{}}MOTION\_DETECTED, NO\_MOTION, \\ UNAVAILABLE, UNKNOWN\end{tabular}       \\ \hline
39 & `entrance\_smart\_lock\_connectivity`   & OFFLINE, ONLINE                                                                                                                                                                                                                                                              & 117 & `office\_smart\_plug`                      & OFF, ON                                                                                             \\ \hline
40 & `entrance\_smart\_lock\_state\_log`     & LOCK\_CMD                                                                                                                                                                                                                                                                    & 118 & `pets`                                     & NONE                                                                                                \\ \hline
41 & `ev\_charger\_schedule`                 & SCHEDULE:23:00-07:00;RATED\_KW:7.2                                                                                                                                                                                                                                           & 119 & `phone\_gps\_tracker`                      & LOCATION:SHANGHAI                                                                                   \\ \hline
42 & `garage\_door\_connectivity`            & OFFLINE, ONLINE                                                                                                                                                                                                                                                              & 120 & `scene\_controller`                        & MOVIE\_MODE, SCENE:GOOD\_NIGHT                                                                      \\ \hline
43 & `garage\_door\_sensor`                  & CLOSED, OPEN, TIMEOUT                                                                                                                                                                                                                                                        & 121 & `security\_camera\_connectivity`           & OFFLINE, ONLINE                                                                                     \\ \hline
44 & `garage\_door\_sensor\_connectivity`    & OFFLINE, ONLINE                                                                                                                                                                                                                                                              & 122 & `shed\_sensor\_connectivity`               & OFFLINE, ONLINE                                                                                     \\ \hline
45 & `garage\_ev\_charger`                   & OFF, ON                                                                                                                                                                                                                                                                      & 123 & `smart\_home\_app`                         & SYNCED                                                                                              \\ \hline
46 & `garage\_light`                         & OFF, ON                                                                                                                                                                                                                                                                      & 124 & `smart\_hub`                               & OFFLINE, ONLINE, RESTART                                                                            \\ \hline
47 & `garage\_motion\_sensor`                & \begin{tabular}[c]{@{}c@{}}MOTION, MOTION\_DETECTED, \\ NO\_MOTION, UNAVAILABLE, UNKNOWN\end{tabular}                                                                                                                                                                        & 125 & `smart\_hub\_command\_log`                 & UNKNOWN\_COMMAND                                                                                    \\ \hline
48 & `garage\_sensor\_connectivity`          & OFFLINE, ONLINE                                                                                                                                                                                                                                                              & 126 & `system\_mode`                             & VACATION\_MODE:ON                                                                                   \\ \hline
49 & `geofence\_event`                       & APPROACHING\_HOME;ETA\_MIN:30;HVAC\_MODE:PREHEAT                                                                                                                                                                                                                             & 127 & `thermostat\_system`                       & MODE:ECO\_AWAY, MODE:LEARNING                                                                       \\ \hline
50 & `guest\_bedroom\_ceiling\_light`        & OFF, ON, TIMEOUT                                                                                                                                                                                                                                                             & 128 & `voice\_assistant`                         & FAILURE\_START:OFFLINE                                                                              \\ \hline
51 & `guest\_bedroom\_motion\_sensor`        & MOTION\_DETECTED, NO\_MOTION                                                                                                                                                                                                                                                 & 129 & `wifi\_router`                             & OFFLINE, ONLINE                                                                                     \\ \hline
52 & `hallway\_light`                        & OFF, ON                                                                                                                                                                                                                                                                      & 130 & `zone\_controller`                         & ZONE\_ACTIVATE:LIVING\_ROOM\_LIGHTS                                                                 \\ \hline
53 & `hallway\_motion\_sensor`               & MOTION, NO\_MOTION                                                                                                                                                                                                                                                           & 131 & `basement\_power\_meter`                   & 2000W                                                                                               \\ \hline
54 & `hallway\_motion\_sensor\_connectivity` & OFFLINE, ONLINE                                                                                                                                                                                                                                                              & 132 & `basement\_thermostat`                     & 15-18°C                                                                                             \\ \hline
55 & `home\_profile`                         & \begin{tabular}[c]{@{}c@{}}NEWBORN:TRUE, PET:CAT, \\ WORK\_SCHEDULE:EARLY\_SHIFT\_STARTED, \\ WORK\_SCHEDULE:NIGHT\_SHIFT\_WORKER, \\ WORK\_SCHEDULE:REMOTE\_FULL\_TIME\end{tabular}                                                                                         & 133 & `bathroom\_power\_meter`                   & 2000W                                                                                               \\ \hline
56 & `hvac\_diagnostics`                     & STRUCTURED                                                                                                                                                                                                                                                                   & 134 & `bathroom\_thermostat`                     & 22-25°C                                                                                             \\ \hline
57 & `hvac\_failure\_prediction`             & RISK\_HIGH                                                                                                                                                                                                                                                                   & 135 & `bedroom\_power\_meter`                    & 64-2000W                                                                                            \\ \hline
58 & `hvac\_history`                         & STRUCTURED                                                                                                                                                                                                                                                                   & 136 & `bedroom\_thermostat`                      & 17.1-24°C                                                                                           \\ \hline
59 & `ip\_camera\_front`                     & FAILURE\_START:OFFLINE                                                                                                                                                                                                                                                       & 137 & `entrance\_door\_angle`                    & 0-69°                                                                                               \\ \hline
60 & `kitchen\_coffee\_maker`                & OFF, ON                                                                                                                                                                                                                                                                      & 138 & `entrance\_door\_sensor\_battery`          & 28-100\%                                                                                            \\ \hline
61 & `kitchen\_dishwasher`                   & OFF, ON                                                                                                                                                                                                                                                                      & 139 & `garage\_thermostat`                       & 18°C                                                                                                \\ \hline
62 & `kitchen\_door\_sensor`                 & CLOSED, OPEN                                                                                                                                                                                                                                                                 & 140 & `grid\_voltage`                            & 206V                                                                                                \\ \hline
63 & `kitchen\_door\_sensor\_connectivity`   & OFFLINE, ONLINE                                                                                                                                                                                                                                                              & 141 & `kitchen\_power\_meter`                    & 0-9000W                                                                                             \\ \hline
64 & `kitchen\_fridge`                       & CLOSED, OPEN                                                                                                                                                                                                                                                                 & 142 & `kitchen\_smart\_plug\_power`              & 0-1582W                                                                                             \\ \hline
65 & `kitchen\_hvac`                         & COOLING, HEATING, OFF                                                                                                                                                                                                                                                        & 143 & `kitchen\_thermostat`                      & 20-25°C                                                                                             \\ \hline
66 & `kitchen\_kettle`                       & OFF, ON                                                                                                                                                                                                                                                                      & 144 & `kitchen\_wifi\_rssi`                      & -85-60DBM                                                                                           \\ \hline
67 & `kitchen\_light`                        & OFF, ON                                                                                                                                                                                                                                                                      & 145 & `living\_room\_light\_power\_meter`        & 60W                                                                                                 \\ \hline
68 & `kitchen\_light\_state\_log`            & PERSISTED:ON                                                                                                                                                                                                                                                                 & 146 & `living\_room\_power\_meter`               & 0-3000W                                                                                             \\ \hline
69 & `kitchen\_microwave`                    & OFF, ON                                                                                                                                                                                                                                                                      & 147 & `living\_room\_smart\_plug\_power`         & 0-162W                                                                                              \\ \hline
70 & `kitchen\_microwave\_state\_log`        & PERSISTED:ON                                                                                                                                                                                                                                                                 & 148 & `living\_room\_thermostat`                 & 18.1-25°C                                                                                           \\ \hline
71 & `kitchen\_motion\_connectivity`         & OFFLINE, ONLINE                                                                                                                                                                                                                                                              & 149 & `living\_room\_tv\_power`                  & 0-180W                                                                                              \\ \hline
72 & `kitchen\_motion\_sensor`               & \begin{tabular}[c]{@{}c@{}}MOTION, MOTION\_DETECTED, \\ NO\_MOTION, TIMEOUT, UNAVAILABLE, \\ UNKNOWN\end{tabular}                                                                                                                                                            & 150 & `living\_room\_window\_opening\_angle`     & 0-42°                                                                                               \\ \hline
73 & `kitchen\_motion\_sensor\_connectivity` & OFFLINE, ONLINE                                                                                                                                                                                                                                                              & 151 & `main\_power\_meter`                       & 120-18000W                                                                                          \\ \hline
74 & `kitchen\_motion\_sensor\_state\_log`   & PERSISTED:NO\_MOTION                                                                                                                                                                                                                                                         & 152 & `master\_bedroom\_smart\_plug\_power`      & 0-118W                                                                                              \\ \hline
75 & `kitchen\_oven\_plug`                   & POWER\_OFF, POWER\_ON                                                                                                                                                                                                                                                        & 153 & `master\_bedroom\_window\_opening\_angle`  & 0-29°                                                                                               \\ \hline
76 & `kitchen\_range\_hood`                  & OFF, ON                                                                                                                                                                                                                                                                      & 154 & `office\_smart\_plug\_power`               & 0-147W                                                                                              \\ \hline
77 & `kitchen\_range\_hood\_state\_log`      & PERSISTED:ON                                                                                                                                                                                                                                                                 & 155 & `smart\_hub\_command\_latency`             & 120-348MS                                                                                           \\ \hline
78 & `kitchen\_smart\_lock`                  & LOCKED, UNLOCKED                                                                                                                                                                                                                                                             &     &                                            &                                                                                                     \\ \hline
\end{tabular}}
\end{table*}

\subsubsection{Context-Independent Anomalies}
\begin{table*}[h]
\centering
\caption{Context-Independent Anomaly Types}
\label{tab:CIanomalytype}
\resizebox{\textwidth}{!}{
\begin{tabular}{|c|c|c|}
\hline
Anomaly Type            & Description                                                                                                                                                                                                                                                                                               & Subtype                                                                                                                                                                                                                                                                                                                                                                                                                                                                                                                     \\ \hline
device\_malfunction     & \begin{tabular}[c]{@{}c@{}}Device or sensor malfunction, offline status, or abnormal \\ reporting, or inconsistencies between “control/display status” \\ and telemetry readings, physical location, or other supporting \\ evidence.\end{tabular}                                                        & \begin{tabular}[c]{@{}c@{}}actuator\_position\_anomaly, alarm\_false\_trigger, camera\_offline, \\ configuration\_conflict, invalid\_categorical\_state, \\ leak\_sensor\_false\_positive, leak\_sensor\_suspicious, \\ lock\_door\_state\_inconsistent, sensor\_unavailable, \\ telemetry\_state\_conflict\end{tabular}                                                                                                                                                                                                    \\ \hline
safety\_hazard          & \begin{tabular}[c]{@{}c@{}}High-risk safety events or protection failures related to \\ smoke, gas, water leakage, door/window security, \\ or surveillance, which may lead to fire, explosion, \\ intrusion, or property loss.\end{tabular}                                                              & \begin{tabular}[c]{@{}c@{}}alarm\_sensor\_state\_conflict, garage\_access\_risk, \\ gas\_detected, open\_entry\_while\_away, \\ security\_disabled\_with\_activity, security\_system\_disabled, \\ security\_system\_disabled\_with\_activity, smoke\_detected, \\ smoke\_false\_alarm\_or\_sensor\_fault, surveillance\_disabled, \\ surveillance\_disabled\_with\_activity, unlocked\_entry\_while\_away, \\ unlocked\_entry\_with\_activity, water\_leak\_detected, \\ window\_left\_open\_in\_bad\_weather\end{tabular} \\ \hline
physical\_violation     & \begin{tabular}[c]{@{}c@{}}Out-of-range, missing, or improperly formatted values, or \\ values that contradict basic physical laws or environmental \\ facts (e.g., humidity readings inconsistent with signs of \\ leakage, or abnormal temperature differences between \\ adjacent rooms).\end{tabular} & \begin{tabular}[c]{@{}c@{}}humidity\_invalid\_value, humidity\_out\_of\_range, \\ humidity\_physics\_conflict, temperature\_invalid\_value, \\ temperature\_out\_of\_range, temperature\_spatial\_inconsistency, \\ temperature\_unrealistic\_indoor, thermal\_distribution\_anomaly\end{tabular}                                                                                                                                                                                                                           \\ \hline
compound\_anomaly       & \begin{tabular}[c]{@{}c@{}}Multiple high-risk signals occur concurrently within the \\ same time window, forming compound risks such as fire, \\ intrusion, or utility-related hazards (with overlapping risks \\ that are harder to interpret).\end{tabular}                                             & \begin{tabular}[c]{@{}c@{}}fire\_signature, intrusion\_signature, multiple\_safety\_alerts, \\ security\_posture\_failure, water\_electric\_risk\_signature\end{tabular}                                                                                                                                                                                                                                                                                                                                                    \\ \hline
environmental\_conflict & \begin{tabular}[c]{@{}c@{}}The on/off status, mode, or target temperature of the HVAC \\ system is inconsistent with ambient conditions such as indoor \\ temperature or open-window ventilation, leading to \\ conflicting control objectives or unreasonable adjustment \\ directions.\end{tabular}     & \begin{tabular}[c]{@{}c@{}}cooling\_target\_conflict, cooling\_temperature\_conflict, \\ heating\_target\_conflict, heating\_temperature\_conflict, \\ hvac\_not\_responding, hvac\_running\_while\_off, \\ hvac\_status\_mode\_mismatch, hvac\_window\_conflict\end{tabular}                                                                                                                                                                                                                                               \\ \hline
occupancy\_mismatch     & \begin{tabular}[c]{@{}c@{}}Occupancy profiling (e.g., at-home or away status) conflicts with \\ activity evidence such as motion sensors, access control logs, \\ or power consumption patterns, indicating potential labeling\\  errors or suggesting abnormal behavior or intrusion risks.\end{tabular} & \begin{tabular}[c]{@{}c@{}}activity\_signature\_with\_no\_occupants, \\ device\_active\_while\_away, motion\_with\_no\_occupants, \\ multi\_room\_activity\_with\_no\_occupants\end{tabular}                                                                                                                                                                                                                                                                                                                                \\ \hline
\end{tabular}}
\end{table*}

Table~\ref{tab:CIanomalytype} presents the anomaly categories and their corresponding subtypes in the context-independent task of \textsf{SmartBench}.

\subsubsection{Context-Dependent Anomalies}
\begin{table*}[h]
\centering
\caption{Context-Dependent Anomaly Types}
\label{tab:CDanomalytype}
\resizebox{\textwidth}{!}{
\begin{tabular}{|c|c|c|}
\hline
Anomaly Type            & Description                                                                                                                                                                                                                                                                                                                                              & Subtype                                                                                                                                                                                                                       \\ \hline
device\_malfunction     & \begin{tabular}[c]{@{}c@{}}Device or sensor failure, disconnection, or performance degradation, \\ manifested as loss of controllability, lack of reporting, \\ or reported readings/states that contradict other sensor data \\ or power consumption patterns observed during the same time period.\end{tabular}                                        & \begin{tabular}[c]{@{}c@{}}device\_failure, parameter\_drift, parameter\_spike, sensor\_stuck, \\ sensor\_drift, thermostat\_failure, device\_aging, \\ component\_cascade\_failure, predictive\_failure\_risk\end{tabular}   \\ \hline
causal\_temporal        & \begin{tabular}[c]{@{}c@{}}Anomalous causal and temporal relationships between \\ events, such as expected events not occurring, events \\ happening out of order, significantly delayed responses,\\  or durations/frequencies deviating from normal patterns.\end{tabular}                                                                             & \begin{tabular}[c]{@{}c@{}}causal\_break, causal\_delay, duration\_anomaly, \\ sequence\_violation, timing\_anomaly, periodicity\_anomaly, \\ frequency\_anomaly, period\_anomaly\end{tabular}                                \\ \hline
statistical\_anomaly    & \begin{tabular}[c]{@{}c@{}}The data significantly deviates from historical patterns \\ or similar scenarios in terms of overall level, fluctuation \\ amplitude, or the synchronization between key indicators.\end{tabular}                                                                                                                             & \begin{tabular}[c]{@{}c@{}}correlation\_anomaly, baseline\_deviation, point\_anomaly, \\ contextual\_anomaly, collective\_anomaly, spurious\_event, \\ distribution\_shift, apparent\_anomaly\_mostly\_seasonal\end{tabular}  \\ \hline
safety\_hazard          & \begin{tabular}[c]{@{}c@{}}High-risk safety incidents related to water leakage, \\ intrusion, security breaches, device compromise, \\ or electricity theft, which may pose direct threats \\ to personal or property safety.\end{tabular}                                                                                                               & \begin{tabular}[c]{@{}c@{}}water\_leak, hidden\_water\_leak, collective\_security\_trigger,  \\ security\_issue, unauthorized\_entry, \\ progressive\_device\_compromise, power\_theft\_upstream\end{tabular}                 \\ \hline
climate\_anomaly        & \begin{tabular}[c]{@{}c@{}}Anomalies in HVAC system cooling/heating/ventilation and \\ zone control, such as malfunctions, overheating, \\ ventilation issues, mode conflicts, or imbalanced zone regulation.\end{tabular}                                                                                                                               & \begin{tabular}[c]{@{}c@{}}hvac\_failure, excess\_heat, ventilation\_anomaly, \\ hvac\_mode\_conflict, hvac\_zone\_conflict, \\ hvac\_zone\_imbalance\end{tabular}                                                            \\ \hline
network\_disconnect     & \begin{tabular}[c]{@{}c@{}}Anomalies in network links, gateways, or integration channels,\\  leading to communication interruptions/fluctuations, \\ state desynchronization, or cross-zone synchronization failures.\end{tabular}                                                                                                                       & \begin{tabular}[c]{@{}c@{}}network\_failure, hub\_offline, wireless\_interference, \\ network\_topology\_anomaly, state\_sync\_failure, \\ zone\_sync\_failure, integration\_failure, \\ sensor\_fusion\_anomaly\end{tabular} \\ \hline
common\_sense\_conflict & \begin{tabular}[c]{@{}c@{}}The event stream exhibits clear inconsistencies between control \\ mode/state flags and actual device status or power consumption \\ changes, or displays phenomena such as frequent short-term \\ device toggling, partial execution of scenes, or long-term \\ failure to achieve intended control objectives.\end{tabular} & \begin{tabular}[c]{@{}c@{}}rule\_conflict, scene\_mode\_conflict, scene\_partial\_failure, \\ coordination\_conflict, feedback\_loop, security\_mode\_conflict, \\ energy\_management\_conflict\end{tabular}                  \\ \hline
occupancy\_anomaly      & \begin{tabular}[c]{@{}c@{}}Occupancy profiles (e.g., home, away, sleep) are inconsistent \\ with behavioral evidence such as actions, access control, \\ or power consumption, or there are anomalous activities\\  such as spontaneous device activation without clear triggers.\end{tabular}                                                           & \begin{tabular}[c]{@{}c@{}}occupant\_incapacitation, unexpected\_activity, \\ occupancy\_contradiction, ghost\_device\_activation\end{tabular}                                                                                \\ \hline
energy\_anomaly         & \begin{tabular}[c]{@{}c@{}}Anomalous fluctuations in power consumption, discrepancies \\ between different meters or sub-metering statistics, or \\ economic anomalies (e.g., cost increases despite expected \\ savings), deviating from normal energy usage patterns.\end{tabular}                                                                     & \begin{tabular}[c]{@{}c@{}}power\_fluctuation, energy\_discrepancy, \\ economic\_anti\_optimization\end{tabular}                                                                                                              \\ \hline
\end{tabular}}
\end{table*}
Table~\ref{tab:CDanomalytype} presents the anomaly categories and their corresponding subtypes in the context-independent task of \textsf{SmartBench}.

% \subsection{Case Study}
% \label{app:casestudy}

\subsection{Limitation}
\label{sec:limitation}
While \textsf{SmartBench} provides valuable support for the development of next-generation smart home assistants, we acknowledge several limitations.
% In the current version of our dataset, each sample assumes that all device actions and states occur within a specified set of physical areas, and that environmental parameters remain consistent across these areas. However,
In real-world scenarios, user routines may include multiple areas over time, and the impact of devices on environmental parameters outside their immediate physical context would be better to be considered on a case-by-case basis.
However, constructing samples that reflect cross-zone effects is particularly challenging. Data generated through synthetic methods often fails to capture the subtle effects of devices on different physical spaces. The most reliable solution would be to build a physical testbed and collect data from volunteers in real environments. However, the cost of such an effort is far beyond the scope of this study. Furthermore, it is valuable to explore more anomaly types. These directions will be part of our future work.

\end{document}